\documentclass{article}

\usepackage{iclr2022_conference,times}
\iclrfinalcopy

\usepackage[utf8]{inputenc} %
\usepackage[T1]{fontenc}    %
\usepackage{doi}
\usepackage{hyperref}       %
\usepackage{url}            %
\usepackage{booktabs}       %
\usepackage{amsfonts}       %
\usepackage{nicefrac}       %
\usepackage{microtype}      %
\usepackage{xcolor}         %

\usepackage{makecell}

\usepackage{amsmath}
\usepackage{amssymb}
\usepackage{hyperref}
\usepackage{graphicx}
\usepackage{caption}
\usepackage{subcaption}
\usepackage{mathtools}
\usepackage{chngcntr}
\usepackage{enumitem}
\usepackage[normalem]{ulem}

\newcommand{\cm}{\,\bar{\ast}\,} %
\newcommand{\N}{\mathcal{N}} 

\renewcommand{\emph}[1]{\textit{#1}}

\newenvironment{my_enumerate}{
\begin{enumerate}[leftmargin=20pt,rightmargin=-5pt, topsep=0pt]
\setlength{\labelindent}{10pt}
\setlength{\itemsep}{0pt}
\setlength{\parskip}{0pt}
\setlength{\parsep}{0pt}}
{\end{enumerate}}

\title{Group equivariant\\ neural posterior estimation}

\author{%
  Maximilian~Dax \\
   Max Planck Institute for Intelligent Systems\\
   T\"ubingen, Germany\\
   \texttt{maximilian.dax@tuebingen.mpg.de} \\
   \And
   Stephen R.~Green \\
   Max Planck Institute for Gravitational Physics \\
   Potsdam, Germany \\
   \texttt{stephen.green@aei.mpg.de}
   \And
   Jonathan~Gair \\
   Max Planck Institute for Gravitational Physics \\
   Potsdam, Germany \\
   \And
   Michael~Deistler \\
   Machine Learning in Science, University of T\"ubingen\\
     T\"ubingen, Germany\\
   \And
   Bernhard~Sch\"olkopf \\
   Max Planck Institute for Intelligent Systems~~~~\\
   T\"ubingen, Germany\\
   \And
   Jakob H.~Macke \\
   Max Planck Institute for Intelligent Systems \&\\
      Machine Learning in Science, University of T\"ubingen\\
  T\"ubingen, Germany\\
   \\
}

\begin{document}

\maketitle
\begin{abstract}

  Simulation-based inference with conditional neural density
  estimators is a powerful approach to solving inverse problems in
  science. However, these methods typically treat the underlying
  forward model as a black box, with no way to exploit geometric
  properties such as equivariances. Equivariances are common
  in scientific models, however integrating them directly into
  expressive inference networks (such as normalizing flows) is not
  straightforward. We here describe an alternative method to
  incorporate equivariances under joint transformations of parameters
  and data. Our method---called group equivariant neural posterior
  estimation (GNPE)---is based on self-consistently standardizing the
  ``pose'' of the data while estimating the posterior over
  parameters. It is architecture-independent, and applies both to
  exact and approximate equivariances. As a real-world application,
  we use GNPE for amortized inference of astrophysical binary black
  hole systems from gravitational-wave observations. We show that GNPE
  achieves state-of-the-art accuracy while reducing inference times by
  three orders of magnitude.
 
\end{abstract}

\section{Introduction}

Bayesian inference provides a means of characterizing a system by
comparing models against data. Given a forward model or likelihood
$p(x|\theta)$ for data $x$ described by parameters $\theta$, and
a prior $p(\theta)$, the Bayesian posterior
is proportional to the product,
$p(\theta|x) \propto p(x|\theta)p(\theta)$. Sampling techniques such
as Markov Chain Monte Carlo (MCMC) can be used to build up a posterior
distribution provided the likelihood and prior can be evaluated.

For models with intractable or expensive likelihoods (as often arise
in scientific applications) simulation-based (or likelihood-free)
inference methods offer a powerful
alternative~\citep{cranmer2020frontier}. In particular, neural
posterior estimation (NPE)~\citep{papamakarios2016fast} uses
expressive conditional density estimators such as normalizing
flows~\citep{rezende2015variational,papamakarios2019normalizing} to
build surrogates for the posterior. These are trained using model
simulations $x\sim p(x|\theta)$, and allow for rapid sampling for any
$x \sim p(x)$, thereby amortizing training costs across future
observations.  NPE and other density-estimation methods for
simulation-based
inference~\citep{gutmann2016bayesian,papamakarios2019sequential,hermans2019likelihood}
have been reported to be more simulation-efficient \citep{lueckmann2021benchmarking} than classical
likelihood-free methods such as Approximate Bayesian
Computation~\citep{sisson2018handbook}.

\begin{figure}
  \centering
  \begin{minipage}[c]{0.4\textwidth}
    \centering
    \includegraphics[trim={10cm 6.7cm 21cm, 6.8cm},clip,width=0.9\textwidth]{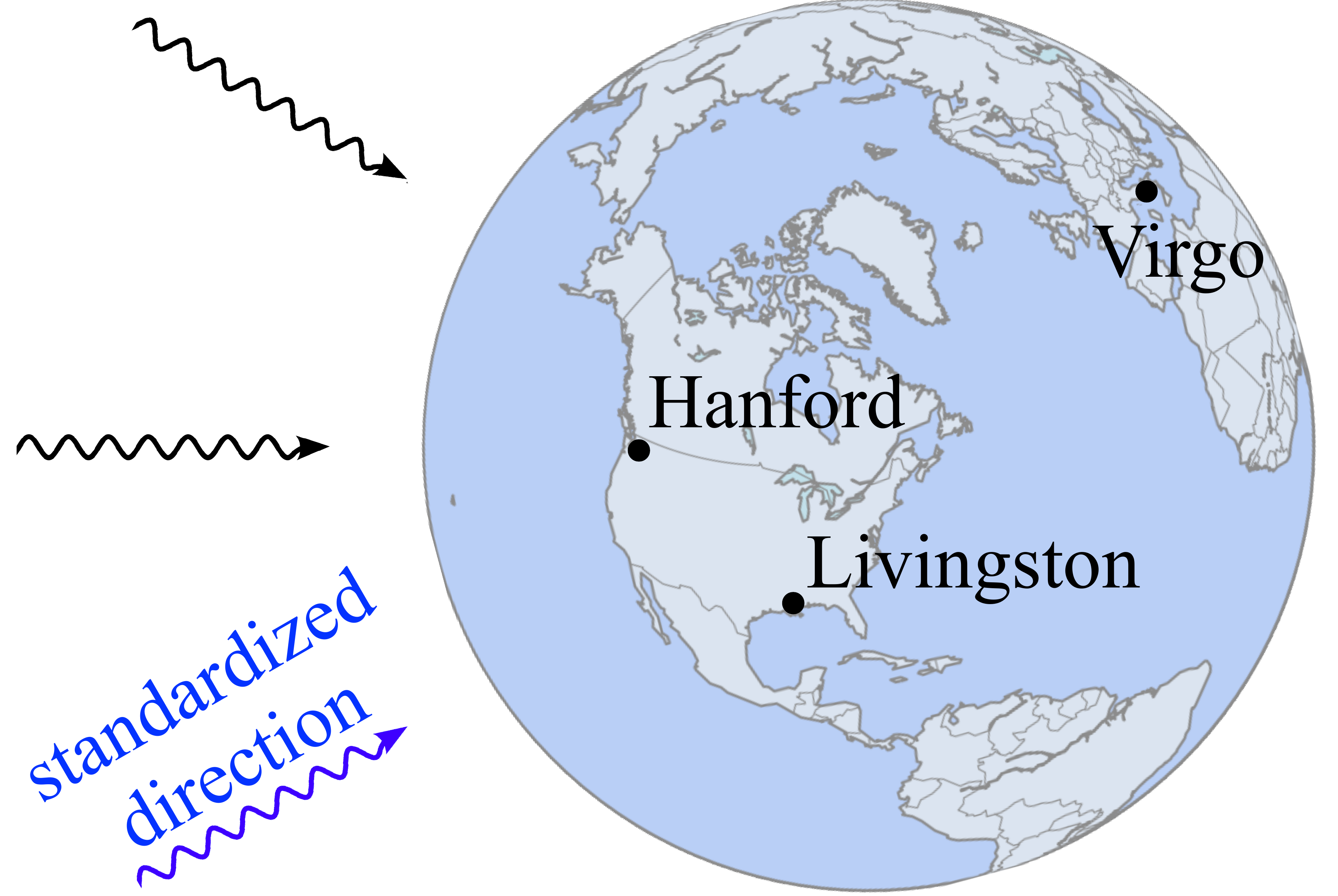}
  \end{minipage}\hfill
  \begin{minipage}[c]{0.57\textwidth}
    \caption{By standardizing the source sky position, a GW signal can
      be made to arrive at the same time in all three LIGO/Virgo
      detectors. However, since this also changes the projection of
      the signal onto the detectors, it defines only an
      \emph{approximate} equivariance. Nevertheless, our proposed GNPE
      algorithm simplifies inference by simultaneously inferring
      \emph{and} standardizing the incident direction.}
    \label{fig:GW-equiv}
  \end{minipage}
\end{figure}

Training an inference network for any $x\sim p(x)$ can nevertheless
present challenges due to the large number of training samples and
powerful networks required. The present study is motivated by the problem of gravitational-wave (GW) data analysis. Here the
task is to infer properties of astrophysical black-hole mergers
based on GW signals observed at the LIGO and Virgo observatories on
Earth. Due to the complexity of signal models, it has previously not
been possible to train networks to estimate posteriors to the same
accuracy as conventional likelihood-based
methods~\citep{Veitch:2014wba,Ashton:2018jfp}. The GW posterior,
however, is equivariant\footnote{In physics, the term ``covariant'' is frequently used instead of ``equivariant''.}
under an overall change in the time of arrival
of the data. It is also \emph{approximately} equivariant under a joint
change in the sky position and (by triangulation) individual shifts in
the arrival times in each detector (Fig.~\ref{fig:GW-equiv}). If we
could constrain these parameters \emph{a priori}, we could therefore
apply time shifts to align the detector data and simplify the
inference task for the remaining parameters.

More generally, we consider forward models 
with known equivariances 
under group transformations
applied jointly to data and parameters. Our aim is to exploit this
knowledge to \emph{standardize the pose of the data}\footnote{We adopt
  the language from computer vision by~\citet{jaderberg2015spatial}.}
and simplify analysis. The obvious roadblock is that the pose is
contained in the set of parameters $\theta$ and is therefore unknown prior to
inference. Here we describe \emph{group equivariant} neural posterior
estimation (GNPE), a method to self-consistently infer parameters
\emph{and} standardize the pose. The basic approach is to introduce a
\emph{proxy} for the pose---a blurred version---on which one
conditions the posterior. The pose of the data is then transformed
based on the proxy, placing it in a band about the standard value, and
resulting in an easier inference task. Finally, the joint posterior
over $\theta$ and the pose proxy can be sampled at inference time
using Gibbs sampling. 

The standard method to incorporating equivariances 
is to integrate them directly into network
architectures, e.g., to use convolutional networks for translational equivariances. Although these approaches can be highly effective, they impose design constraints on
 network architectures. For GWs, for example, we use specialized
embedding networks 
to extract signal waveforms
from frequency-domain data, as well as expressive normalizing flows to
estimate the posterior---neither of which is straightforward to make
explicitly equivariant. We also have complex equivariance connections
between subsets of parameters and data, including approximate
equivariances. The GNPE algorithm is extremely general: it is
architecture-independent, it applies whether equivariances are exact
or approximate, and it allows for arbitrary equivariance relations
between parameters and data.

We discuss related work in Sec.~\ref{sec:related-work} and describe the GNPE algorithm in Sec.~\ref{sec:methods}. In
Sec.~\ref{sec:toy-example} we apply GNPE to a toy example with exact
translational equivariance, showing comparable simulation efficiency to NPE combined with a convolutional network. In Sec.~\ref{sec:gwpe} we
show that standard NPE does not achieve adequate accuracy for GW parameter inference, even with an essentially
unlimited number of simulations. In contrast, GNPE achieves highly accurate posteriors
at a computational cost three orders of magnitude lower than bespoke MCMC approaches~\citep{Veitch:2014wba}. The present paper describes the GNPE method which we developed for GW analysis~\citep{dax2021real}, and extends it to general equivariance transformations which makes it applicable to a wide range of problems. A detailed description of GW results is presented in~\citet{dax2021real}.

\section{Related work}
\label{sec:related-work}

The most common way of integrating equivariances into
machine learning algorithms is to use equivariant network
architectures~\citep{krizhevsky2012imagenet,cohen2016group}. 
This can be in conflict with design considerations such as data
representation and flexibility of the architecture, and imposes
constraints such as locality. GNPE achieves complete separation of
equivariances from these considerations, requiring only the
ability to efficiently transform the pose. 

Normalizing flows are particularly well
suited to NPE, and there has been significant progress in
constructing equivariant flows \citep{boyda2021sampling}. However, 
these studies consider joint transformations of parameters of the
base space and sample space---\emph{not} joint
transformation of data and parameters for
\emph{conditional} flows, as we consider here.

GNPE enables end-to-end equivariances from data to parameters. Consider, by contrast, a conditional normalizing flow with a convolutional embedding network: the equivariance
persists through the embedding network but is broken by the flow. Although this may improve learning, it does not enforce an end-to-end
equivariance. This contrasts with an \emph{invariance}, for which the above would be sufficient. 
Finally, GNPE can also be applied if the 
equivariance is only \emph{approximate}.

Several other approaches integrate domain knowledge of
the forward model~\citep{baydin2019etalumis,brehmer2020mining} by
considering a ``gray-box'' setting. GNPE allows us to incorporate
high-level domain knowledge about approximate equivariances of forward
models without requiring access to its implementation or internal
states of the simulator. Rather, it can be applied to ``black-box''
code.

An alternative approach to incorporate geometrical knowledge into
classical
likelihood-free inference algorithms (e.g., Approximate Bayesian
Computation, see \citep{sisson2018handbook}) is by constructing
\citep{fearnhead2012constructing} or learning
\citep{jiang2017learning,chen2021neural} equivariant summary
statistics $s(x)$, which are used as input to the inference algorithm
instead of the raw data $x$. However, designing equivariant summary
statistics (rather than invariant ones) can be challenging, and
furthermore inference will be biased if the equivariance only holds approximately.

Past studies using machine-learning techniques for amortized GW parameter
inference~\citep{Gabbard:2019rde,Chua:2019wwt,Green:2020dnx,delaunoy2020lightning}
all consider simplified problems (e.g., only a subset of
parameters, a simplified posterior, or a limited treatment of detector noise). In contrast, the GNPE-based study in \citet{dax2021real}
is the only one to treat the full amortized parameter inference problem with  accuracy matching standard methods.

\section{Methods}
\label{sec:methods}

\subsection{Neural posterior estimation}

NPE~\citep{papamakarios2016fast,greenberg2019automatic} is a simulation-based inference method that 
directly targets the posterior. Given a dataset of prior parameter samples
$\theta^{(i)}\sim p(\theta)$ and corresponding model simulations
$x^{(i)}\sim p(x|\theta^{(i)})$, it trains a neural density estimator
$q(\theta|x)$ to estimate $p(\theta|x)$. This is achieved by
minimizing the loss
\begin{equation}\label{eq:L-npe}
  \mathcal L_{\text{NPE}} = \mathbb{E}_{p(\theta)}\mathbb{E}_{p(x|\theta)}\left[ - \log q(\theta | x) \right]
\end{equation}
across the dataset of
$(\theta^{(i)}, x^{(i)})$ pairs. This maximum
likelihood objective leads to 
recovery of $p(\theta|x)$ if $q(\theta|x)$ is sufficiently flexible. 
Normalizing flows~\citep{rezende2015variational,durkan2019neural} are a particularly expressive class of conditional density estimators commonly used for NPE.

NPE amortizes inference: once $q(\theta|x)$ is trained, inference is
very fast for any observed data $x_{\text{o}}$, so training costs are
shared across observations. The approach is also extremely
flexible, as it treats the forward model as a black box, relying only on prior
samples and model simulations. In many situations,
however, these data have known structure that one wants to
exploit to improve learning.

\subsection{Equivariances under transformation groups}

In this work we describe a generic method to incorporate equivariances
under joint transformations of $\theta$ and $x$ into NPE. A typical
example arises when inferring the position of an object from image
data. In this case, if we spatially translate an image $x$ by some
offset $\vec d$---effected by \emph{relabeling the pixels}---then the inferred position
$\theta$ should also transform by $\vec d$---by \emph{addition} to the position coordinates $\theta$. 
Translations are composable and
invertible, and there exists a trivial identity translation, so the
set of translations has a natural group structure. Our method works
for any continuous transformation group, including rotations,
dilations, etc., and in this section we keep the discussion general.

For a transformation group $G$, we denote the action of $g\in G$ on
parameters and data as
\begin{align}
  \theta &\to g\theta,\\
  x &\to T_gx.
\end{align}
Here, $T_g$ refers to the group representation under which the data
transform (e.g., for image translations, the pixel relabeling). We
adopt the natural convention that $G$ is defined by its action on
$\theta$, so we do not introduce an explicit representation on
parameters. The posterior distribution $p(\theta|x)$ is said to be
\emph{equivariant} under $G$ if, when the parameter and data
spaces are jointly $G$-transformed, the posterior is unchanged, i.e.,
\begin{equation}\label{eq:equiv-posterior}
  p(\theta | x) = p(g\theta|T_gx) |\det J_g|, \qquad \forall g\in G.
\end{equation}
The right-hand side comes from the change-of-variables rule. 
For translations the Jacobian $J_g$ has unit determinant, but we
include it for generality. For NPE, we are concerned with equivariant
posteriors, however it is often more natural to think of equivariant
forward models (or likelihoods). An equivariant likelihood and an
\emph{invariant} prior together yield an equivariant posterior
(App.~\ref{sec:appendix-equivariance-relations}).

Our goal is to use equivariances to simplify the data---to
$G$-transform $x$ such that $\theta$ is taken to a fiducial value. For
the image example, this could mean translating the object of interest
to the center. In general, $\theta$ can also include parameters
unchanged under $G$ (e.g., the color of the object), so we denote the
corresponding standardized parameters by $\theta_0$. These are related
to $\theta$ by a group transformation denoted $g^\theta$, such that
$g^\theta\theta_0 = \theta$. We refer to $g^\theta$ as the ``pose'' of
$\theta$, and standardizing the pose means to take it to the group
identity element $e\in G$. Applying $T_{(g^\theta)^{-1}}$ to the data
space effectively reduces its dimensionality, making it easier to
interpret for a neural network.

Although the preceding discussion applies to equivariances that hold
exactly, our method in fact generalizes to \emph{approximate}
equivariances. We say that a posterior is approximately
equivariant under $G$ if \eqref{eq:equiv-posterior} does \emph{not}
hold, but standardizing the pose nevertheless reduces the effective
dimensionality of the dataset. An approximately equivariant posterior
can arise if an exact equivariance of the forward model is broken by a
non-invariant prior, or if the forward model is itself
non-equivariant. 

\subsection{Group equivariant neural posterior estimation}
\label{subsec:gnpe}

We are now presented with the basic problem that we resolve in this
work: how to simultaneously infer the pose of a signal and use that
inferred pose to standardize (or align) the data so as to simplify the
analysis. This is a circular problem because one cannot standardize
the pose (contained in model parameters $\theta$) without first inferring
the pose from the data; and conversely one cannot easily infer the
pose without first simplifying the data by standardizing the pose. 

Our resolution is to start with a rough estimate of the pose, and
iteratively (1) transform the data based on a pose estimate, and (2)
estimate the pose based on the transformed data. To do so, we expand
the parameter space to include \emph{approximate} pose parameters
$\hat g \in G$. These ``pose proxies'' are defined using a kernel to
blur the true pose, i.e., $\hat g = g^\theta\epsilon$ for
$\epsilon \sim \kappa(\epsilon)$; then
$p(\hat g|\theta) = \kappa\left((g^\theta)^{-1}\hat g\right)$. The
kernel $\kappa(\epsilon)$ is a distribution over group elements, which
should be chosen to be concentrated around $e$; we furthermore choose
it to be symmetric. Natural choices for $\kappa(\epsilon)$ include
Gaussian and uniform distributions. For translations, the pose proxy is
simply the true position with additive noise.

Consider now the posterior distribution $p(\theta, \hat g | x)$
over the expanded parameter space. Our iterative algorithm comes from
Gibbs sampling this distribution~\citep{roberts1994simple}
(Fig.~\ref{fig:pose-proxy-illustration}), i.e., alternately
sampling $\theta$ and $\hat g$, conditional on the other parameter and
$x$,
\begin{align}
  \label{eq:gibbs-step1} \theta &\sim p(\theta | x, \hat g),\\
  \hat g &\sim p(\hat g | x, \theta).
\end{align}
The second step just amounts to blurring the pose, since
$p(\hat g| x, \theta) = p(\hat g|\theta) =
\kappa\left((g^\theta)^{-1}\hat g\right)$. The key first step uses a
neural density estimator $q$ that is trained taking advantage of a
standardized pose. 

For an \textbf{equivariant} posterior, the distribution
\eqref{eq:gibbs-step1} can be rewritten as
(App.~\ref{sec:appendix-equivariance-extended-posterior})
\begin{align}\label{eq:alignment}
  p(\theta | x, \hat g) &= p\left(\hat g^{-1}\theta|T_{\hat g^{-1}}x, \hat g^{-1}\hat g\right) \left|\det J_{\hat g}^{-1}\right| 
  \equiv p(\theta' | x') \left|\det J_{\hat g}^{-1}\right|.
\end{align}
For the last equality we defined $\theta' \equiv \hat g^{-1}\theta$ and
$x' \equiv T_{\hat g^{-1}}x$, and we dropped the constant argument
$\hat g^{-1}\hat g = e$. 
This expresses $p(\theta|x,\hat g)$ in terms of the $\hat g$-standardized data $x'$---which is much easier to estimate. 
We train a neural density estimator
$q(\theta'|x')$ to approximate this, by minimizing the loss,
\begin{equation}
  \mathcal{L}_{\text{GNPE}} = \mathbb{E}_{p(\theta)} \mathbb{E}_{p(x|\theta)} \mathbb{E}_{p(\hat g|\theta)} \left[ - \log q\left(\hat g^{-1} \theta  | T_{\hat g^{-1}} x \right) \right]. 
\end{equation}
With a trained $q(\theta'|x')$,
\begin{equation}\label{eq:gnpe-exact-equiv}
  \theta \sim p(\theta | x, \hat g)  \qquad \Longleftrightarrow \qquad \theta = \hat g \theta', \quad \theta' \sim q(\theta' | T_{\hat g^{-1}} x).
\end{equation}
The estimated posterior is equivariant by construction
(App.~\ref{sec:appendix-exact}).

For an \textbf{approximately-equivariant} posterior,
\eqref{eq:gibbs-step1} cannot be transformed to be independent of
$\hat g$. We are nevertheless able to use the conditioning on $\hat g$
to approximately align $x$. 
We therefore
train a neural density estimator $q(\theta|x', \hat g)$, by minimizing
the loss
\begin{equation}
  \mathcal{L}_{\text{GNPE}} = \mathbb{E}_{p(\theta)} \mathbb{E}_{p(x|\theta)} \mathbb{E}_{p(\hat g|\theta)} \left[ - \log q\left(\theta  | T_{\hat g^{-1}} x, \hat g \right) \right]. 
\end{equation}
In general, one may have a combination of exact and
approximate equivariances (see, e.g., Sec.~\ref{sec:gwpe}).

\subsection{Gibbs convergence}

The Gibbs-sampling procedure constructs a Markov chain with
equilibrium distribution $p(\theta, \hat g|x)$. For convergence, the
chain must be transient, aperiodic and
irreducible~\citep{roberts1994simple,gelman2013bayesian}. For sensible
choices of $\kappa(\epsilon)$ the chain is transient and
aperiodic by construction. Further, irreducibility means that the
entire posterior can be reached starting from any point, which should
be possible even for disconnected posteriors provided the kernel is
sufficiently broad.  In general, burn-in truncation and thinning of
the chain is required to ensure (approximately) independent samples.
By marginalizing over $\hat g$ (i.e., ignoring it) we obtain samples
from the posterior $p(\theta|x)$, as desired.\footnote{In practice,
  this results only in \emph{approximate} samples due to the
  asymptotic behaviour of Gibbs sampling, and a potential mismatch
  between the trained $q$ and the targeted true posterior.}

\begin{figure}
    \centering
    \begin{minipage}[c]{0.48\textwidth}
    \includegraphics[trim={0cm 20cm 37cm 0cm},clip,width=\textwidth]{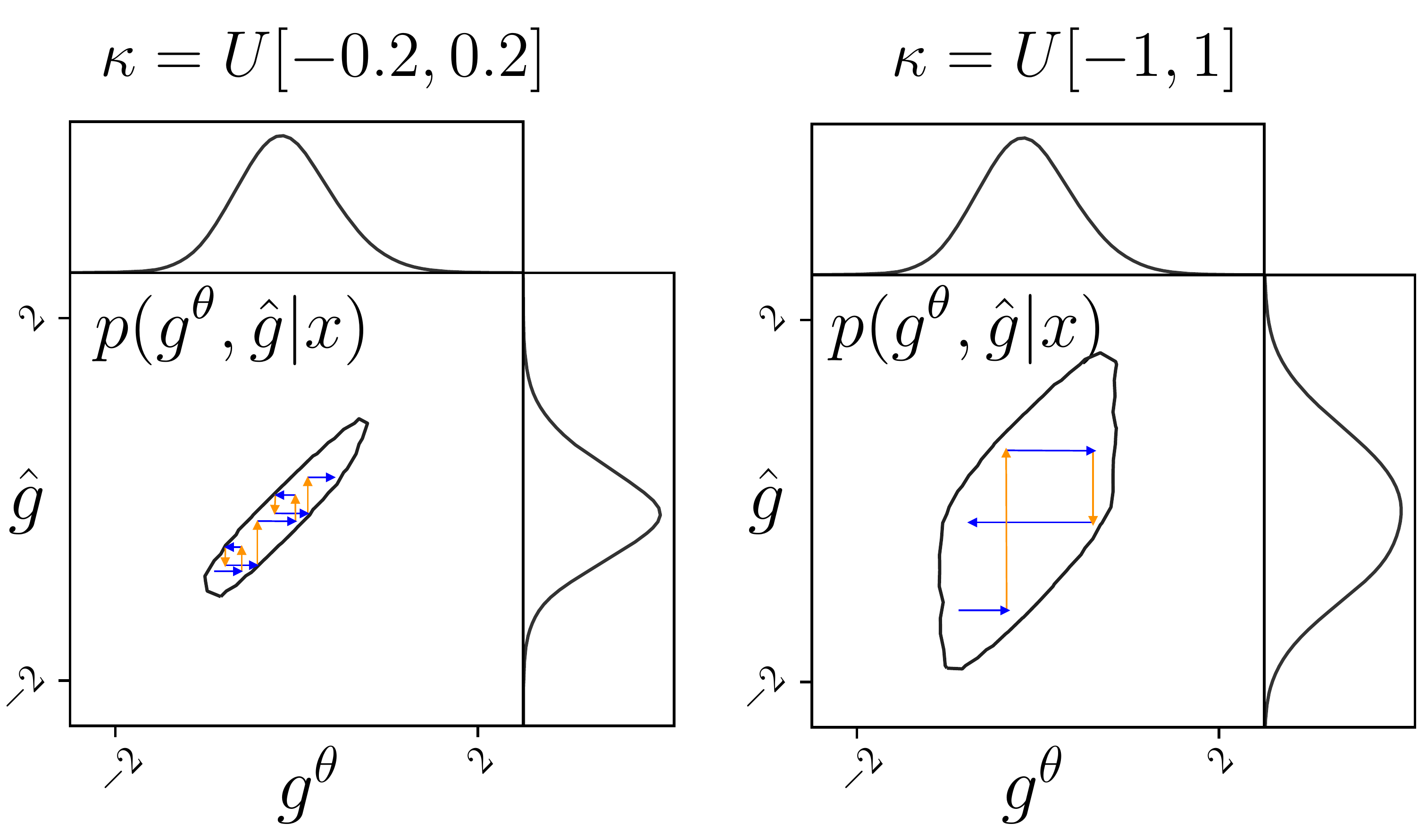}
    \end{minipage}\hfill
    \begin{minipage}[c]{0.50\textwidth}
    \caption{
    We infer $p(g^\theta,\hat g|x)$ with Gibbs sampling by alternately sampling (1) $g^\theta\sim p(g^\theta|x,\hat g)$ (blue) and (2) $\hat g\sim p(\hat g|x, g^\theta)$ (orange). For (1) we use a density estimator $q(g^\theta|T_{\hat g^{-1}}(x),\hat g)$, for (2) the definition $\hat g=g^\theta+\epsilon,\,\epsilon\sim\kappa(\epsilon)$. 
    Pose standardization with $T_{\hat g^{-1}}$ is only allowed due to conditioning on $\hat g$. Increasing the width of $\kappa$ accelerates convergence (due to larger steps in parameter space), at the cost of $\hat g$ being a worse approximation for $g^\theta$, and therefore pose alignment being less effective.
    }
    \label{fig:pose-proxy-illustration}
    \end{minipage}
\end{figure}

Convergence of the chain also informs our choice of
$\kappa(\epsilon)$. For wide $\kappa(\epsilon)$, only a few Gibbs
iterations are needed to traverse the joint posterior
$p(\theta, \hat g | x)$, whereas for narrow $\kappa(\epsilon)$ many
steps are required (Fig.~\ref{fig:pose-proxy-illustration}). In the limiting case of $\kappa(\epsilon)$ a delta distribution (i.e., no blurring) the chain does not deviate from its initial position and therefore fails to converge.\footnote{This also explains why introducing the pose proxy is needed at all: GNPE would not work without it!}  Conversely, a narrower
$\kappa(\epsilon)$ better constrains the pose, which improves the accuracy of the density estimator.
The width of $\kappa$ 
should be chosen based on this practical trade-off between speed and
accuracy; the standard deviation of a typical pose posterior is
usually a good starting point.

In practice, we obtain $N$ samples in parallel by constructing an
ensemble of $N$ Markov chains. We initialize these using samples from
a second neural density estimator $q_\text{init}(g^\theta|x)$, trained
using standard NPE. Gibbs sampling yields a sequence of sample sets
$\{\theta_j^{(i)}\}_{i=1}^N$, $j=0,1,2,\ldots$, each of which
represents a distribution $Q_j(\theta|x)$ over parameters. Assuming a
perfectly trained network, one iteration applied to sample set $j$
yields an updated distribution,
\begin{equation}\label{eq:convergence}
  Q_{j+1}(\theta|x) = p(\theta|x)\left[ \frac{Q_j(\cdot|x) \cm \kappa}{p(\cdot|x) \cm \kappa} \ast \kappa\right](g^\theta).
\end{equation}
The ``$\ast$'' symbol denotes group convolution and ``$\cm$'' the
combination of marginalization and group convolution (see App. \ref{sec:appendix-iterative-inference} for details). 
The true posterior $p(\theta|x)$ is clearly a fixed point of this
sequence, with the number of iterations to convergence determined by
$\kappa$ and the accuracy of the initialization network
$q_\text{init}$. 

\section{Toy example: damped harmonic oscillator}
\label{sec:toy-example}

We now apply GNPE to invert a simple model of a
damped harmonic oscillator. The forward model gives the time-dependent
position $x$ of the oscillator, conditional on its real frequency
$\omega_0$, damping ratio $\beta$, and time of excitation
$\tau$. The time series $x$ is therefore a damped sinusoid starting at
$\tau$ (and zero before). Noise is introduced via a
normally-distributed perturbation of the parameters
$\theta=(\omega_0,\beta,\tau)$, resulting in a  Gaussian
posterior $p(\theta|x)$ (further details in  App.~\ref{sec:appendix-toy-example-forward-model}). The model is constructed such that the
posterior is equivariant under translations in $\tau$,
\begin{equation}
  p(\omega_0,\beta, \tau+\Delta\tau|T_{\Delta\tau}x) = p(\omega_0,\beta, \tau|x),
\end{equation}
so we take $\tau$ to be the pose. 
The equivariance of this model is exact, but it could easily be made approximate by, e.g., introducing $\tau$-dependent noise. 
The prior $p(\tau)$
extends from $-5$~s to $0$~s, so for NPE, the density estimator must learn to
 interpret data from oscillators excited throughout this
range.

\begin{figure}
    \centering
    \includegraphics[trim={0 0 0 0},clip,width=\textwidth]{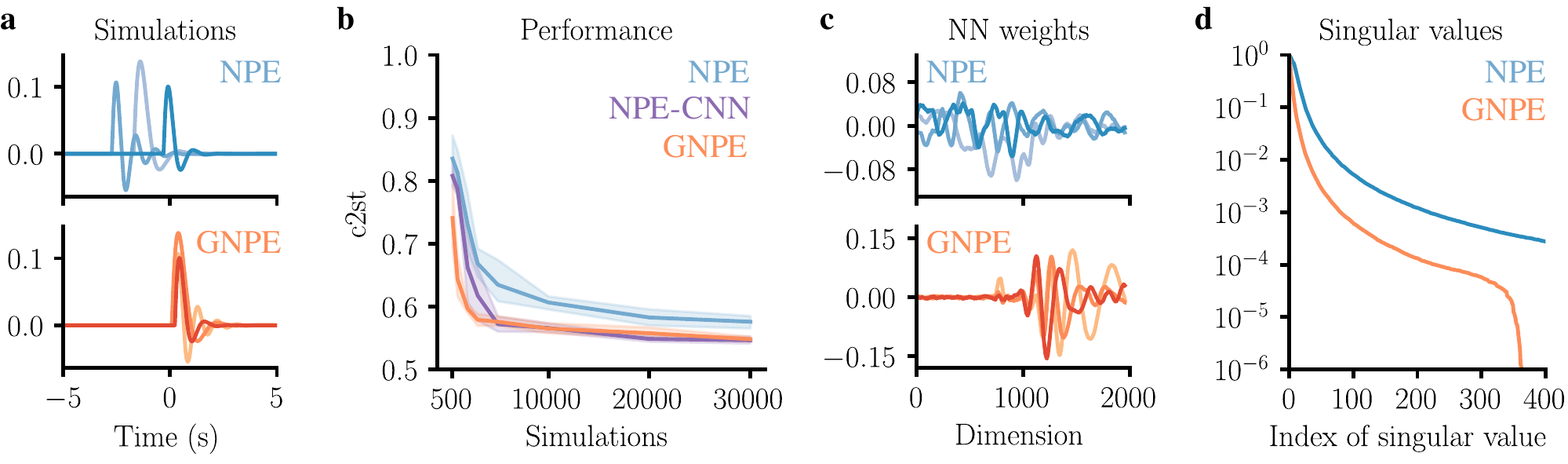}
    \caption{
    Comparison of standard NPE (blue) and GNPE (orange) for the damped harmonic oscillator. 
    {\bf a)} Three sample inputs to the neural density estimators showing GNPE data are pose-standardized. 
    {\bf b)} c2st score performance (best: $0.5$, worst: $1.0$): GNPE significantly outperforms equivariance-agnostic NPE, and is on par with NPE with a convolutional embedding network (purple). 
    {\bf c) } Example filters from the first linear layer of fully trained networks. GNPE filters more clearly capture oscillatory modes. 
    {\bf d)} Singular values of the training data.  Inputs to GNPE $x'$ have smaller effective dimension than raw inputs $x$ to NPE. 
    }
    \label{fig:toy-model-results}
\end{figure}

For GNPE, we shift the data to align the pose near $\tau=0$ using a Gaussian kernel $\kappa=\mathcal{N}[0,(0.1~\text{s})^2]$ (Fig.~\ref{fig:toy-model-results}a).
We then train a neural
density estimator $q(\theta'|x')$ to approximate $p(\theta'|x')$,
where $\theta'\equiv(\omega_0,\beta,-\epsilon)$ and
$x'\equiv T_{-(\tau + \epsilon)}x$ are pose-standardized. We take
$q(\theta'|x')$ to be diagonal Gaussian, matching the known form of
the posterior.
For each experiment, we train until the validation loss stops decreasing. We also train a neural density estimator $q_\text{init}(\tau|x)$ with standard NPE on the same dataset to generate initial GNPE seeds.
To generate $N$ posterior samples we proceed as follows:
\begin{my_enumerate}
    \item Sample $\tau^{(i)}\sim q_\text{init}(\tau|x)$, $i=1,\ldots,N$;
    \item Sample $\epsilon^{(i)} \sim \kappa(\epsilon)$, set $\hat\tau^{(i)}=\tau^{(i)}+\epsilon^{(i)}$, and time-translate the data, $x^{\prime {(i)}} = T_{-\hat\tau^{(i)}}x$;
    \item Sample $\theta^{\prime (i)} \sim q(\theta'|x^{\prime (i)})$, and undo the time translation $\hat\tau^{(i)}$ to obtain $\theta^{(i)}$.
\end{my_enumerate}
We repeat steps 2 and 3 until the distribution over $\tau$
converges. For this toy example and our choice of $\kappa$ only one
iteration is required. For further details of the implementation see
App.~\ref{sec:appendix-toy-example-implementation}.

We evaluate GNPE on five simulations by comparing inferred samples to
 ground-truth posteriors using the c2st
score~\citep{friedman2004multivariate,lopez-paz2018revisiting}. 
This corresponds to the
test accuracy of a classifier trained to discriminate samples from the
target and inferred distributions, and ranges from $0.5$ (best) to
$1.0$ (worst). As baselines we evaluate standard NPE (i) with a
network architecture identical to GNPE and (ii) with a convolutional
embedding network (NPE-CNN; see App.~\ref{sec:appendix-toy-example-implementation}). 
Both approaches that leverage the equivariance, GNPE (by standardizing the pose) and
NPE-CNN (by using a translation-equivariant embedding network),
perform similarly well and far outperform standard NPE (Fig.~\ref{fig:toy-model-results}b). This
underscores the importance of equivariance awareness.
The fact that the NPE
network is trained to interpret signals from oscillators excited at
arbitrary $\tau$, whereas GNPE focuses on signals starting around
$\tau=0$ (up to a small $\epsilon$ perturbation) is also reflected
in in simplified filters in the first layer of the network (Fig.~\ref{fig:toy-model-results}c) and a reduced effective dimension of the input data to GNPE (Fig.~\ref{fig:toy-model-results}d).

\section{Gravitational-wave parameter inference}
\label{sec:gwpe}

Gravitational waves---propagating ripples of space and time---were
first detected in 2015, from the inspiral, merger, and ringdown of a
pair of black holes~\citep{LIGOScientific:2016aoc}. Since that time,
the two LIGO detectors (Hanford and
Livingston)~\citep{TheLIGOScientific:2014jea} as well as the Virgo
detector~\citep{TheVirgo:2014hva} have observed signals from over 50
coalescences of compact binaries involving either black holes or
neutron
stars~\citep{LIGOScientific:2018mvr,LIGOScientific:2020ibl,LIGOScientific:2021usb}. Key scientific results from these observations have included measurements of the properties of stellar-origin black holes that have provided new insights into their origin and evolution~\citep{LIGOScientific:2020kqk}; an independent measurement of the local expansion rate of the Universe, the Hubble constant~\citep{LIGOScientific:2017adf}; and new constraints on the properties of gravity and matter under extreme conditions~\citep{LIGOScientific:2018cki,LIGOScientific:2020tif}.

Quasicircular binary black hole (BBH) systems are characterized by 15
parameters $\theta$, including the component masses and spins, as well
as the space-time position and orientation of the system
(Tab.~\ref{tab:GW-parameters-with-priors}). Given these parameters,
Einstein's theory of general relativity predicts the motion and
emitted gravitational radiation of the binary. The GWs propagate
across billions of light-years to Earth, where they produce a
time-series signal $h_I(\theta)$ in each of the LIGO/Virgo
interferometers $I=\mathrm{H,L,V}$. The signals on Earth are very weak
and embedded in detector noise $n_I$. In part to have a tractable likelihood, the noise is approximated as additive and
stationary Gaussian. The signal and noise models give rise to a
likelihood $p(x|\theta)$ for observed data
$x = \{h_I(\theta) + n_I\}_{I=\mathrm{H,L,V}}$.

Once the LIGO/Virgo detection pipelines are triggered, classical
stochastic samplers are typically employed to determine the parameters
of the progenitor system using Bayesian
inference~\citep{Veitch:2014wba,Ashton:2018jfp}. However, these
methods require millions of likelihood evaluations (and hence
expensive waveform simulations) for each event analyzed. Even using
fast waveform models, it can take $O(\mathrm{day})$ to analyze a
single BBH. Faster inference methods are therefore highly desirable to
cope with growing event rates, more realistic (and expensive) waveform
models, and to make rapid localization predictions for possible
multimessenger counterparts.  Rapid amortized methods such as NPE have
the potential to transform GW data analysis. However, due to the
complexity and high dimensionality\footnote{In our work, we analyze
  8~s data segments between 20~Hz and 1024~Hz. Including also noise
  information, this results in 24,099 input dimensions per detector.}
of GW data, it has been a challenge~\citep{Gabbard:2019rde,Chua:2019wwt,Green:2020dnx,delaunoy2020lightning}
to obtain results of comparable accuracy and
completeness to classical samplers. We now show how GNPE can be used
to exploit equivariances to greatly simplify the inference problem and
achieve for the first time performance indistinguishable from ``ground
truth'' stochastic samplers---at drastically reduced inference times.

\subsection{Equivariances of sky position and coalescence time}
\label{subsec:gwpe-equivariance}

We consider the analysis of BBH systems. Included among the parameters
$\theta$ are the time of coalescence $t_\text{c}$ (as measured at
geocenter) and the sky position (right ascension $\alpha$, declination
$\delta$). Since GWs propagate at the speed of light, these are
related to the times of arrival $t_I$ of the signal in each of the
interferometers.\footnote{We consider observations made in either $n_I=2$ or $3$ interferometers.} Our priors (based on the precision of
detection pipelines) constrain $t_I$ to a range of $\approx 20$~ms,
which is much wider than typical posteriors. Standard NPE inference
networks must therefore be trained on simulations with substantial
time shifts.

The detector coalescence times $t_I$---equivalently,
$(t_c, \alpha, \delta)$---can alternatively be interpreted as the
pose of the data, and standardized using GNPE. The group $G$
transforming the pose factorizes into a direct product of absolute and
relative time shifts,
\begin{equation}
    G = G_\text{abs} \times G_\text{rel}.
\end{equation}
Group elements $g_\text{abs} \in G_\text{abs}$ act by uniform
translation of all $t_I$, whereas $g_\text{rel}\in G_\text{rel}$ act
by individual translation of $t_\text{L}$ and $t_\text{V}$. We work
with data in frequency domain, where time translations act by
multiplication, $T_gx_I = e^{-2\pi i f \Delta t_I}x_I$. Absolute time
shifts correspond to a shift in $t_c$, and are an \emph{exact}
equivariance of the posterior,
$p(g_\text{abs}\theta|T_{g_\text{abs}}x) = p(\theta|x)$. 
Relative time shifts correspond to a change in $(\alpha,\delta)$ (as well as
$t_c$). This is only an \emph{approximate} equivariance, since a change in sky position changes the projection of the incident signal onto the detector arms, leading to a subdominant change to the signal morphology in each detector.

\subsection{Application of GNPE}
\label{sec:gwpe-gnpe}

We use GNPE to standardize the pose within a band around $t_I=0$. We
consider two modes defined by different uniform blurring kernels. The
``accurate'' mode uses a narrow kernel
$\kappa_\text{narrow} = U[-1~\mathrm{ms},1~\mathrm{ms}]^{n_I}$, whereas
the ``fast'' mode uses a wide kernel
$\kappa_\text{wide} = U[-3~\mathrm{ms},3~\mathrm{ms}]^{n_I}$. The latter is intended to converge in just one GNPE iteration, at the cost of having to interpret a wider range of data.

We define the blurred pose proxy $\hat g_I \equiv t_I + \epsilon_I$,
where $\epsilon_I \sim \kappa(\epsilon_I)$. We then train a
conditional density estimator $q(\theta'|x',\hat g_\text{rel})$,
where $\theta' = \hat g_\text{abs}^{-1}\theta$ and
$x' = T_{\hat g^{-1}}x$. That is, we condition $q$ on the relative
time shift (since this is an approximate equivariance) and we
translate parameters by the absolute time shift (since this is an
exact equivariance). We always transform the data by the full time
shift. We train a density estimator
$q_\text{init}(\{t_I\}_{I=\mathrm{H,L,V}}|x)$ using standard NPE to
infer initial pose estimates. 

The difficulty of the inference problem (high data dimensionality, significant noise levels, complex forward model) combined with high accuracy requirements to be scientifically useful requires careful design decisions. 
In particular, we initialize the first layer of the embedding network with principal components of clean waveforms to provide an inductive bias to extract useful information. We further use an expressive neural-spline normalizing flow~\citep{durkan2019neural} to model the complicated GW posterior structure. See App.~\ref{sec:appendix-GW-network-architecture} for details of network architecture and training.

\subsection{Results}
\label{sec:gwpe-results}

\begin{figure}
    \centering
    \begin{minipage}[c]{0.60\textwidth}
    \includegraphics[trim={0cm 0cm 0cm 0cm},clip,width=\textwidth]{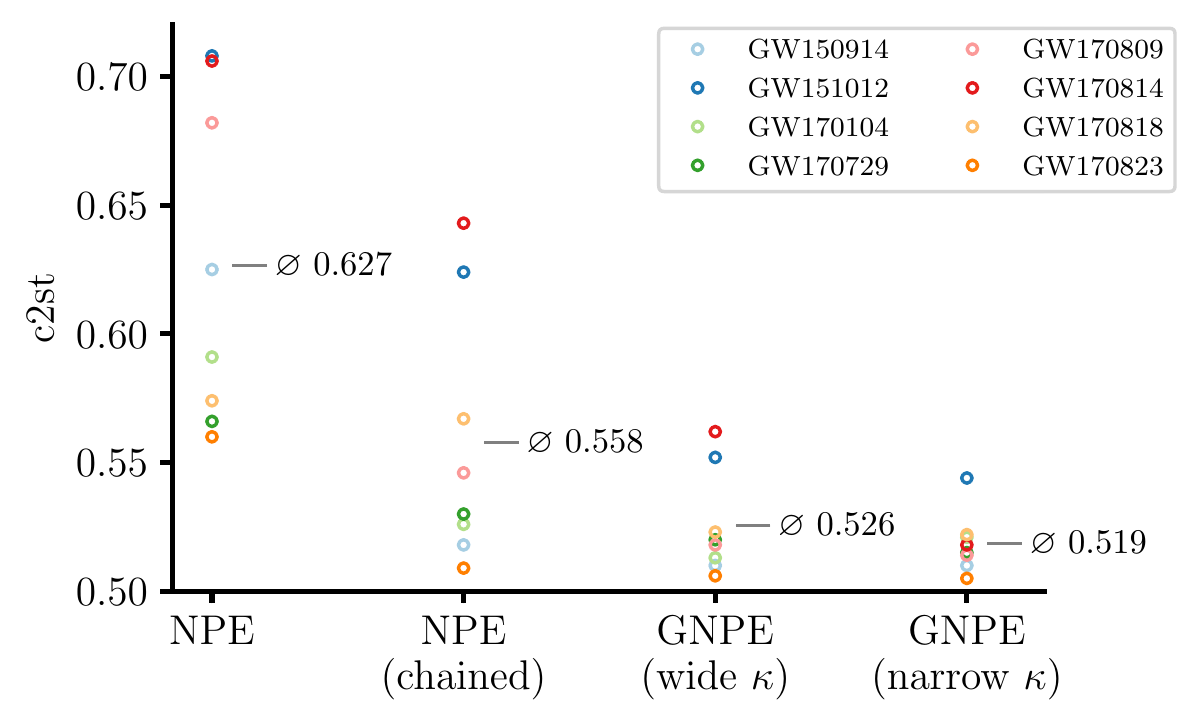}
    \end{minipage}\hfill
    \begin{minipage}[c]{0.36\textwidth}
    \caption{Comparison of estimated posteriors against \textsc{LALInference} MCMC for eight GW events, as quantified by c2st  (best: $0.50$, worst: $1.00$). 
    GNPE with a wide kernel outperforms both NPE baselines, while being only marginally slower (1 iteration $\sim 2$~s). 
    With a narrow kernel and 30 iterations ($\sim 60$~s), we achieve c2st~$<0.55$ across all events. 
    $\varnothing$ indicates the average across all eight events. 
    For an alternative metric (MSE) see Fig.~\ref{fig:GW-ablation-studies-mse}.
    }
    \label{fig:GW-ablation-studies-c2st}
    \end{minipage}
\end{figure}

We evaluate performance on all eight BBH events from the first
Gravitational-Wave Transient Catalog~\citep{LIGOScientific:2018mvr} consistent with our prior (component masses greater than 10~M${}_\odot$). We generate
reference posteriors with the LIGO/Virgo MCMC code
\textsc{LALInference}~\citep{Veitch:2014wba}. We quantify the deviation between NPE samples
and the reference samples using c2st.

We compare performance against two baselines, standard NPE and a modified approach that partially standardizes the pose (``chained NPE''). For the latter, we use the chain rule to decompose the posterior,
\begin{equation}
\label{eq:chained-npe}
    p(\theta|x) = p(\phi,\lambda|x) = p(\phi|x,\lambda)\cdot p(\lambda|x),
\end{equation}
where $\lambda=(t_c,\alpha,\delta)$ are the pose parameters and $\phi\subset\theta$ collects the remaining 12
parameters. We use standard NPE to train a flow $q(\lambda|x)$ to
estimate $p(\lambda|x)$, and a flow $q(\phi|x',\lambda)$ to estimate
$p(\phi|x,\lambda)$. The latter flow is
conditioned on $\lambda$, which we use to standardize the pose of
$x$. 
In contrast to GNPE, this 
approach is sensitive to the initial pose estimate
$q(\lambda|x)$, which limits accuracy
(Figs.~\ref{fig:GW-ablation-studies-c2st} and~\ref{fig:GW-ablation-studies-GW170814-pose-and-proxy}). 
We note that all
hyperparameters of the flow and training protocol (see
App.~\ref{sec:appendix-GW-network-architecture}) were
extensively optimized on NPE, and then transferred to GNPE without
modification, resulting in conservative estimates of the performance
advantage of GNPE. Fast-mode GNPE converges in one
iteration, whereas accurate-mode requires 30 
(convergence is assessed by the JS divergence between the inferred pose posteriors from two successive iterations). 

Standard NPE performs well on some GW events but lacks the required
accuracy for most of them, with c2st scores up to $0.71$
(Fig.~\ref{fig:GW-ablation-studies-c2st}). Chained NPE performs better across the dataset, but
performs poorly on events such as GW170814, for which the initial pose
estimate is inaccurate.  
Indeed, we find that inaccuracies of that baseline can be almost entirely attributed to the initial pose estimate (Fig.~\ref{fig:GW-ablation-studies-GW170814-chained-pose}). 
Fast-mode GNPE with only one iteration is already more robust to this effect due to the blurring operation of the pose proxy (Fig.~\ref{fig:GW-ablation-studies-GW170814-pose-and-proxy}). 
Both GNPE models significantly outperform
the baselines, with accurate-mode obtaining c2st scores $<0.55$
across all eight events. We emphasize that the c2st score is sensitive
to any deviation between 
reference samples and
samples from the inferred posterior. On a recent benchmark
by~\citet{lueckmann2021benchmarking} on examples with much
lower parameter \emph{and} data dimensions, even state-of-the-art
SBI algorithms rarely reached c2st scores below $0.6$. The fact that
GNPE achieves scores around $0.52$---i.e., posteriors which are nearly
indistinguishable from the reference---on this challenging,
high-dimensional, real-world example underscores the power of
exploiting equivariances with GNPE. 

\begin{figure}
    \centering
    \begin{subfigure}{0.48\textwidth}
        \includegraphics[width=\textwidth]{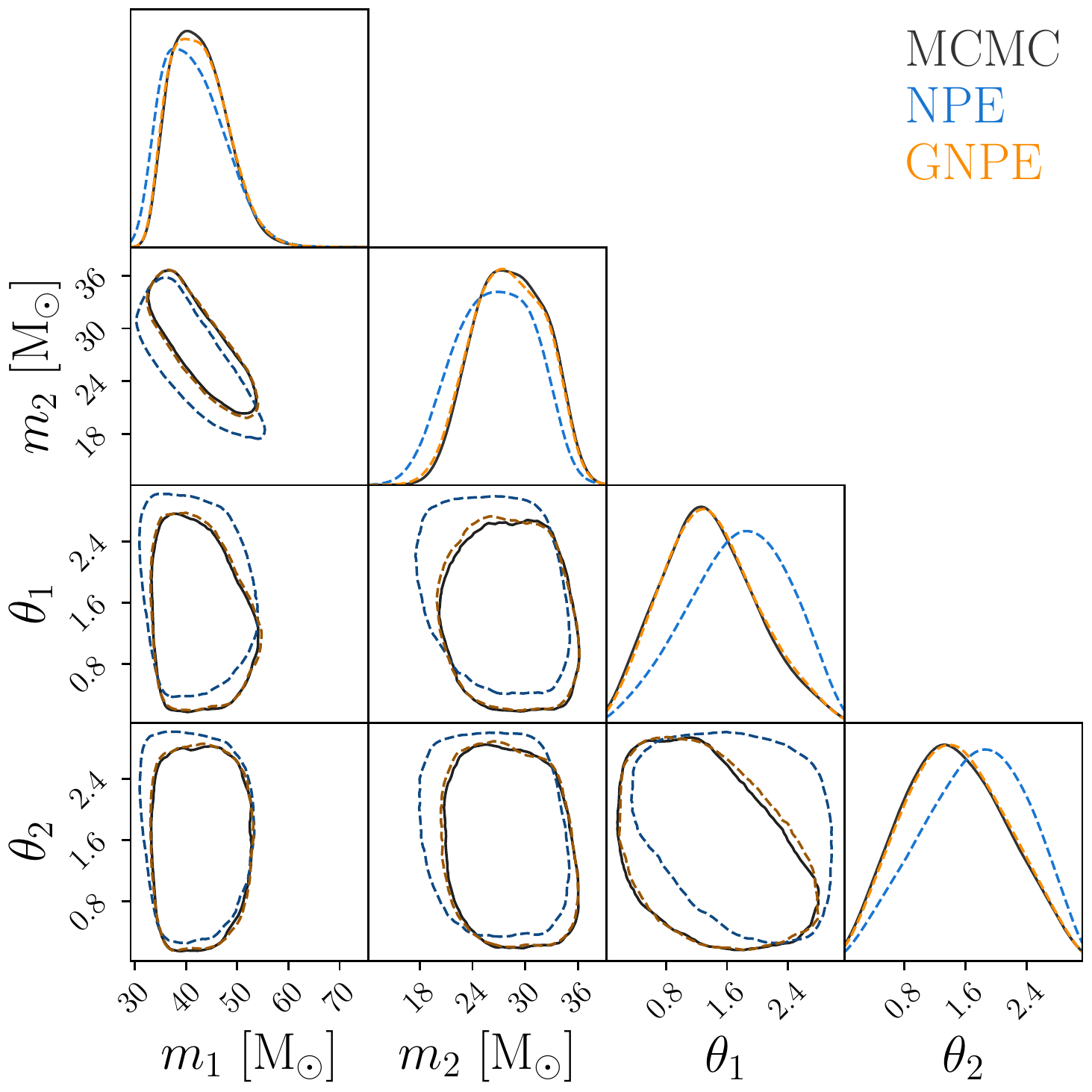}
    \end{subfigure}
    \hfill
    \begin{subfigure}{0.48\textwidth}
        \includegraphics[width=\textwidth]{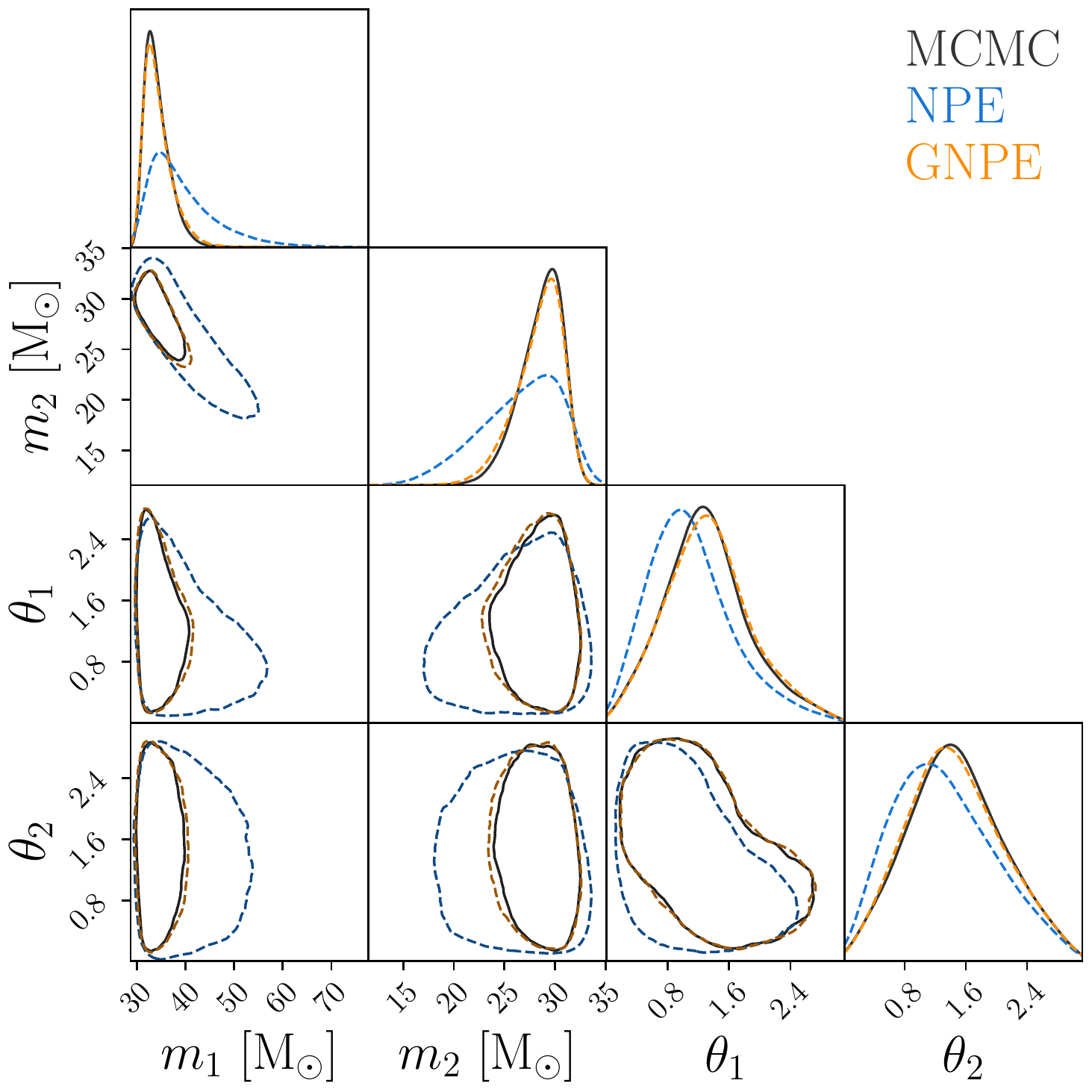}
    \end{subfigure}
    \caption{Corner plots for the GW events GW170809 (left) and
      GW170814 (right), plotting 1D marginals on the diagonal and 90\%
      credible regions for the 2D correlations. We display the two
      black hole masses $m_1$ and $m_2$ and two spin parameters
      $\theta_1$ and $\theta_2$ (note that the full posterior is
      15-dimensional). NPE does not accurately reproduce the MCMC
      posterior, while accurate-mode GNPE matches the MCMC results well. 
      For a plot with all baselines see Fig.~\ref{fig:GW-corner-ablation-studies_all}.
      }
    \label{fig:GW-corner-ablation-studies}
\end{figure}

Finally, we visualize posteriors for two events, GW170809 and GW170814, in Fig.~\ref{fig:GW-corner-ablation-studies}. The quantitative
agreement between GNPE and MCMC 
(Fig.~\ref{fig:GW-ablation-studies-c2st}) is visible from the
overlapping marginals for all parameters displayed. NPE, by contrast, deviates significantly from MCMC in
terms of shape and position. Note that we show a failure case of NPE
here; for other events, such as GW170823, deviations of NPE from
the reference posterior are less clearly visible.

\section{Conclusions}
\label{sec:conclusion}

We described GNPE, an approach to incorporate exact---and even
\emph{approximate}---equivariances under joint transformations of data
and parameters into simulation-based inference. GNPE can be applied to
black-box scientific forward models and any inference network
architecture. 
It requires similar training
times compared to NPE, while the added complexity at inference time
depends on the number of GNPE iterations (adjustable, but typically
$O(10)$). We show with two examples that exploiting equivariances with
GNPE can yield large gains in simulation efficiency and accuracy.

For the motivating problem of GW parameter estimation, GNPE achieves
for the first time rapid amortized inference with results virtually
indistinguishable from MCMC~\citep{dax2021real}. This is an extremely
challenging ``real-world'' scientific problem, with high-dimensional
input data, complex signals, and significant noise levels. It combines
exact and approximate equivariances, and there is no clear path to
success without their inclusion along with GW-specialized
architectures and expressive density estimators.

\subsubsection*{Ethics Statement}
Our method is primarily targeted at scientific applications, and we do not foresee direct applications which are ethically problematic.
In the context of GW analysis, we hope that GNPE contributes to reducing the required amount of compute, in particular when the rate of detections increases with more sensitive detectors in the future.

\subsubsection*{Reproducibility Statement}
The experimental setup for the toy model is described in App.~\ref{sec:appendix-toy-example-implementation}. We also provide the code at \href{https://tinyurl.com/wmbjajv8}{https://tinyurl.com/wmbjajv8}. 
The setup for GW parameter inference is described in App.~\ref{sec:appendix-GW-network-architecture}. The experiments in section~\ref{sec:gwpe} can be reproduced using the \textsc{Dingo} code \href{https://github.com/dingo-gw/dingo}{https://github.com/dingo-gw/dingo}.

\subsubsection*{Acknowledgments}
We thank A. Buonanno, T. Gebhard, J.M. L\"uckmann, S. Ossokine, M. P\"urrer and C. Simpson for helpful discussions. We thank the anonymous reviewer for coming up with the illustration of GNPE in App.~\ref{sec:app-Gaussian-likelihood-prior}. This research has made use of data, software and/or web tools obtained from the Gravitational Wave Open Science Center (https://www.gw-openscience.org/), a service of LIGO Laboratory, the LIGO Scientific Collaboration and the Virgo Collaboration. LIGO Laboratory and Advanced LIGO are funded by the United States National Science Foundation (NSF) as well as the Science and Technology Facilities Council (STFC) of the United Kingdom, the Max-Planck-Society (MPS), and the State of Niedersachsen/Germany for support of the construction of Advanced LIGO and construction and operation of the GEO600 detector. Additional support for Advanced LIGO was provided by the Australian Research Council. Virgo is funded, through the European Gravitational Observatory (EGO), by the French Centre National de Recherche Scientifique (CNRS), the Italian Istituto Nazionale di Fisica Nucleare (INFN) and the Dutch Nikhef, with contributions by institutions from Belgium, Germany, Greece, Hungary, Ireland, Japan, Monaco, Poland, Portugal, Spain. This material is based upon work supported by NSF’s LIGO Laboratory which is a major facility fully funded by the National Science Foundation. 
M. Dax thanks the Hector Fellow Academy for support. M. Deistler thanks the International Max Planck Research School for Intelligent Systems (IMPRS-IS) for support. B.S. and J.H.M. are members of the MLCoE, EXC number 2064/1 – Project number 390727645. This work was supported by the German Federal Ministry of Education and Research (BMBF): Tübingen AI Center, FKZ: 01IS18039A. 
We use \verb|PyTorch|~\citep{pytorch}, \verb|nflows|~\citep{nflows} and \verb|sbi|~\citep{Tejero:2020sbi} for the implementation of our neural networks. The plots are generated with \verb|matplotlib|~\citep{matplotlib} and \verb|ChainConsumer|~\citep{chainconsumer}.

\bibliographystyle{iclr2022_conference}
\bibliography{mybib_arxiv.bib}

\begin{thebibliography}{50}
\providecommand{\natexlab}[1]{#1}
\providecommand{\url}[1]{\texttt{#1}}
\expandafter\ifx\csname urlstyle\endcsname\relax
  \providecommand{\doi}[1]{doi: #1}\else
  \providecommand{\doi}{doi: \begingroup \urlstyle{rm}\Url}\fi

\bibitem[Aasi et~al.(2015)]{TheLIGOScientific:2014jea}
J.~Aasi et~al.
\newblock {Advanced LIGO}.
\newblock \emph{Class. Quant. Grav.}, 32:\penalty0 074001, 2015.
\newblock \doi{10.1088/0264-9381/32/7/074001}.

\bibitem[Abbott et~al.(2016)]{LIGOScientific:2016aoc}
B.~P. Abbott et~al.
\newblock {Observation of Gravitational Waves from a Binary Black Hole Merger}.
\newblock \emph{Phys. Rev. Lett.}, 116\penalty0 (6):\penalty0 061102, 2016.
\newblock \doi{10.1103/PhysRevLett.116.061102}.

\bibitem[Abbott et~al.(2017)]{LIGOScientific:2017adf}
B.~P. Abbott et~al.
\newblock {A gravitational-wave standard siren measurement of the Hubble
  constant}.
\newblock \emph{Nature}, 551\penalty0 (7678):\penalty0 85--88, 2017.
\newblock \doi{10.1038/nature24471}.

\bibitem[Abbott et~al.(2018)]{LIGOScientific:2018cki}
B.~P. Abbott et~al.
\newblock {GW170817: Measurements of neutron star radii and equation of state}.
\newblock \emph{Phys. Rev. Lett.}, 121\penalty0 (16):\penalty0 161101, 2018.
\newblock \doi{10.1103/PhysRevLett.121.161101}.

\bibitem[Abbott et~al.(2019)]{LIGOScientific:2018mvr}
B.~P. Abbott et~al.
\newblock {GWTC-1: A Gravitational-Wave Transient Catalog of Compact Binary
  Mergers Observed by LIGO and Virgo during the First and Second Observing
  Runs}.
\newblock \emph{Phys. Rev. X}, 9\penalty0 (3):\penalty0 031040, 2019.
\newblock \doi{10.1103/PhysRevX.9.031040}.

\bibitem[Abbott et~al.(2021{\natexlab{a}})]{LIGOScientific:2020ibl}
R.~Abbott et~al.
\newblock {GWTC-2: Compact Binary Coalescences Observed by LIGO and Virgo
  During the First Half of the Third Observing Run}.
\newblock \emph{Phys. Rev. X}, 11:\penalty0 021053, 2021{\natexlab{a}}.
\newblock \doi{10.1103/PhysRevX.11.021053}.

\bibitem[Abbott et~al.(2021{\natexlab{b}})]{LIGOScientific:2020kqk}
R.~Abbott et~al.
\newblock {Population Properties of Compact Objects from the Second LIGO-Virgo
  Gravitational-Wave Transient Catalog}.
\newblock \emph{Astrophys. J. Lett.}, 913\penalty0 (1):\penalty0 L7,
  2021{\natexlab{b}}.
\newblock \doi{10.3847/2041-8213/abe949}.

\bibitem[Abbott et~al.(2021{\natexlab{c}})]{LIGOScientific:2020tif}
R.~Abbott et~al.
\newblock {Tests of general relativity with binary black holes from the second
  LIGO-Virgo gravitational-wave transient catalog}.
\newblock \emph{Phys. Rev. D}, 103\penalty0 (12):\penalty0 122002,
  2021{\natexlab{c}}.
\newblock \doi{10.1103/PhysRevD.103.122002}.

\bibitem[Abbott et~al.(2021{\natexlab{d}})]{LIGOScientific:2021usb}
R.~Abbott et~al.
\newblock {GWTC-2.1: Deep Extended Catalog of Compact Binary Coalescences
  Observed by LIGO and Virgo During the First Half of the Third Observing Run}.
\newblock 8 2021{\natexlab{d}}.

\bibitem[Acernese et~al.(2015)]{TheVirgo:2014hva}
F.~Acernese et~al.
\newblock {Advanced Virgo: a second-generation interferometric gravitational
  wave detector}.
\newblock \emph{Class. Quant. Grav.}, 32\penalty0 (2):\penalty0 024001, 2015.
\newblock \doi{10.1088/0264-9381/32/2/024001}.

\bibitem[Ashton et~al.(2019)]{Ashton:2018jfp}
Gregory Ashton et~al.
\newblock {BILBY: A user-friendly Bayesian inference library for
  gravitational-wave astronomy}.
\newblock \emph{Astrophys. J. Suppl.}, 241\penalty0 (2):\penalty0 27, 2019.
\newblock \doi{10.3847/1538-4365/ab06fc}.

\bibitem[Baydin et~al.(2019)Baydin, Shao, Bhimji, Heinrich, Meadows, Liu, Munk,
  Naderiparizi, Gram-Hansen, Louppe, et~al.]{baydin2019etalumis}
Atilim~G{\"u}ne{\c{s}} Baydin, Lei Shao, Wahid Bhimji, Lukas Heinrich, Lawrence
  Meadows, Jialin Liu, Andreas Munk, Saeid Naderiparizi, Bradley Gram-Hansen,
  Gilles Louppe, et~al.
\newblock Etalumis: Bringing probabilistic programming to scientific simulators
  at scale.
\newblock In \emph{Proceedings of the international conference for high
  performance computing, networking, storage and analysis}, pp.\  1--24, 2019.

\bibitem[Boh\'e et~al.(2016)Boh\'e, Hannam, Husa, Ohme, P\"urrer, and
  Schmidt]{Bohe:2016}
Alejandro Boh\'e, Mark Hannam, Sascha Husa, Frank Ohme, Michael P\"urrer, and
  Patricia Schmidt.
\newblock {PhenomPv2 -- technical notes for the LAL implementation}.
\newblock \emph{LIGO Technical Document, LIGO-T1500602-v4}, 2016.
\newblock URL \url{https://dcc.ligo.org/LIGO-T1500602/public}.

\bibitem[Boyda et~al.(2021)Boyda, Kanwar, Racani{\`e}re, Rezende, Albergo,
  Cranmer, Hackett, and Shanahan]{boyda2021sampling}
Denis Boyda, Gurtej Kanwar, S{\'e}bastien Racani{\`e}re, Danilo~Jimenez
  Rezende, Michael~S Albergo, Kyle Cranmer, Daniel~C Hackett, and Phiala~E
  Shanahan.
\newblock Sampling using su (n) gauge equivariant flows.
\newblock \emph{Physical Review D}, 103\penalty0 (7):\penalty0 074504, 2021.

\bibitem[Brehmer et~al.(2020)Brehmer, Louppe, Pavez, and
  Cranmer]{brehmer2020mining}
Johann Brehmer, Gilles Louppe, Juan Pavez, and Kyle Cranmer.
\newblock Mining gold from implicit models to improve likelihood-free
  inference.
\newblock \emph{Proceedings of the National Academy of Sciences}, 117\penalty0
  (10):\penalty0 5242--5249, 2020.

\bibitem[Chen et~al.(2021)Chen, Zhang, Gutmann, Courville, and
  Zhu]{chen2021neural}
Yanzhi Chen, Dinghuai Zhang, Michael~U Gutmann, Aaron Courville, and Zhanxing
  Zhu.
\newblock Neural approximate sufficient statistics for implicit models.
\newblock In \emph{Ninth International Conference on Learning Representations
  2021}, 2021.

\bibitem[Chua \& Vallisneri(2020)Chua and Vallisneri]{Chua:2019wwt}
Alvin J.~K. Chua and Michele Vallisneri.
\newblock {Learning Bayesian posteriors with neural networks for
  gravitational-wave inference}.
\newblock \emph{Phys. Rev. Lett.}, 124\penalty0 (4):\penalty0 041102, 2020.
\newblock \doi{10.1103/PhysRevLett.124.041102}.

\bibitem[Cohen \& Welling(2016)Cohen and Welling]{cohen2016group}
Taco Cohen and Max Welling.
\newblock Group equivariant convolutional networks.
\newblock In \emph{International conference on machine learning}, pp.\
  2990--2999. PMLR, 2016.

\bibitem[Cranmer et~al.(2020)Cranmer, Brehmer, and Louppe]{cranmer2020frontier}
Kyle Cranmer, Johann Brehmer, and Gilles Louppe.
\newblock The frontier of simulation-based inference.
\newblock \emph{Proceedings of the National Academy of Sciences}, 117\penalty0
  (48):\penalty0 30055--30062, 2020.

\bibitem[Dax et~al.(2021)Dax, Green, Gair, Macke, Buonanno, and
  Sch\"olkopf]{dax2021real}
Maximilian Dax, Stephen~R. Green, Jonathan Gair, Jakob~H. Macke, Alessandra
  Buonanno, and Bernhard Sch\"olkopf.
\newblock {Real-Time Gravitational Wave Science with Neural Posterior
  Estimation}.
\newblock \emph{Phys. Rev. Lett.}, 127\penalty0 (24):\penalty0 241103, 2021.
\newblock \doi{10.1103/PhysRevLett.127.241103}.

\bibitem[Delaunoy et~al.(2020)Delaunoy, Wehenkel, Hinderer, Nissanke, Weniger,
  Williamson, and Louppe]{delaunoy2020lightning}
Arnaud Delaunoy, Antoine Wehenkel, Tanja Hinderer, Samaya Nissanke, Christoph
  Weniger, Andrew~R Williamson, and Gilles Louppe.
\newblock Lightning-fast gravitational wave parameter inference through neural
  amortization.
\newblock \emph{Third Workshop on Machine Learning and the Physical Sciences
  (NeurIPS 2020)}, 2020.

\bibitem[Durkan et~al.(2019)Durkan, Bekasov, Murray, and
  Papamakarios]{durkan2019neural}
Conor Durkan, Artur Bekasov, Iain Murray, and George Papamakarios.
\newblock Neural spline flows.
\newblock \emph{Advances in Neural Information Processing Systems},
  32:\penalty0 7511--7522, 2019.

\bibitem[Durkan et~al.(2020)Durkan, Bekasov, Murray, and Papamakarios]{nflows}
Conor Durkan, Artur Bekasov, Iain Murray, and George Papamakarios.
\newblock {nflows}: normalizing flows in {PyTorch}, November 2020.
\newblock URL \url{https://doi.org/10.5281/zenodo.4296287}.

\bibitem[Farr et~al.(2014)Farr, Ochsner, Farr, and O'Shaughnessy]{Farr:2014qka}
Benjamin Farr, Evan Ochsner, Will~M. Farr, and Richard O'Shaughnessy.
\newblock {A more effective coordinate system for parameter estimation of
  precessing compact binaries from gravitational waves}.
\newblock \emph{Phys. Rev. D}, 90\penalty0 (2):\penalty0 024018, 2014.
\newblock \doi{10.1103/PhysRevD.90.024018}.

\bibitem[Fearnhead \& Prangle(2012)Fearnhead and
  Prangle]{fearnhead2012constructing}
Paul Fearnhead and Dennis Prangle.
\newblock Constructing summary statistics for approximate bayesian computation:
  semi-automatic approximate bayesian computation.
\newblock \emph{Journal of the Royal Statistical Society: Series B (Statistical
  Methodology)}, 74\penalty0 (3):\penalty0 419--474, 2012.

\bibitem[Friedman(2004)]{friedman2004multivariate}
Jerome Friedman.
\newblock On multivariate goodness-of-fit and two-sample testing.
\newblock In \emph{Conference on Statistical Problems in Particle Physics,
  Astrophysics and Cosmology}, 2004.

\bibitem[Gabbard et~al.(2019)Gabbard, Messenger, Heng, Tonolini, and
  Murray-Smith]{Gabbard:2019rde}
Hunter Gabbard, Chris Messenger, Ik~Siong Heng, Francesco Tonolini, and
  Roderick Murray-Smith.
\newblock {Bayesian parameter estimation using conditional variational
  autoencoders for gravitational-wave astronomy}, 2019.

\bibitem[Gelman et~al.(2013)Gelman, Carlin, Stern, Dunson, Vehtari, and
  Rubin]{gelman2013bayesian}
Andrew Gelman, John~B Carlin, Hal~S Stern, David~B Dunson, Aki Vehtari, and
  Donald~B Rubin.
\newblock \emph{Bayesian data analysis}.
\newblock CRC press, 2013.

\bibitem[Green \& Gair(2021)Green and Gair]{Green:2020dnx}
Stephen~R. Green and Jonathan Gair.
\newblock {Complete parameter inference for GW150914 using deep learning}.
\newblock \emph{Mach. Learn. Sci. Tech.}, 2\penalty0 (3):\penalty0 03LT01,
  2021.
\newblock \doi{10.1088/2632-2153/abfaed}.

\bibitem[Greenberg et~al.(2019)Greenberg, Nonnenmacher, and
  Macke]{greenberg2019automatic}
David Greenberg, Marcel Nonnenmacher, and Jakob Macke.
\newblock Automatic posterior transformation for likelihood-free inference.
\newblock In \emph{International Conference on Machine Learning}, pp.\
  2404--2414. PMLR, 2019.

\bibitem[Gutmann \& Corander(2016)Gutmann and Corander]{gutmann2016bayesian}
Michael~U Gutmann and Jukka Corander.
\newblock Bayesian optimization for likelihood-free inference of
  simulator-based statistical models.
\newblock \emph{The Journal of Machine Learning Research}, 17\penalty0
  (1):\penalty0 4256--4302, 2016.

\bibitem[Hannam et~al.(2014)Hannam, Schmidt, Bohé, Haegel, Husa, Ohme,
  Pratten, and Pürrer]{Hannam:2013oca}
Mark Hannam, Patricia Schmidt, Alejandro Bohé, Leïla Haegel, Sascha Husa,
  Frank Ohme, Geraint Pratten, and Michael Pürrer.
\newblock {Simple Model of Complete Precessing Black-Hole-Binary Gravitational
  Waveforms}.
\newblock \emph{Phys. Rev. Lett.}, 113\penalty0 (15):\penalty0 151101, 2014.
\newblock \doi{10.1103/PhysRevLett.113.151101}.

\bibitem[Hermans et~al.(2020)Hermans, Begy, and Louppe]{hermans2019likelihood}
Joeri Hermans, Volodimir Begy, and Gilles Louppe.
\newblock Likelihood-free mcmc with approximate likelihood ratios.
\newblock In \emph{Proceedings of the 37th International Conference on Machine
  Learning}, volume~98 of \emph{Proceedings of Machine Learning Research}.
  PMLR, 2020.

\bibitem[{Hinton}(2016)]{chainconsumer}
S.~R. {Hinton}.
\newblock {ChainConsumer}.
\newblock \emph{The Journal of Open Source Software}, 1:\penalty0 00045, August
  2016.
\newblock \doi{10.21105/joss.00045}.

\bibitem[Hunter(2007)]{matplotlib}
J.~D. Hunter.
\newblock Matplotlib: A 2d graphics environment.
\newblock \emph{Computing in Science \& Engineering}, 9\penalty0 (3):\penalty0
  90--95, 2007.
\newblock \doi{10.1109/MCSE.2007.55}.

\bibitem[Jaderberg et~al.(2015)Jaderberg, Simonyan, Zisserman,
  et~al.]{jaderberg2015spatial}
Max Jaderberg, Karen Simonyan, Andrew Zisserman, et~al.
\newblock Spatial transformer networks.
\newblock \emph{Advances in neural information processing systems},
  28:\penalty0 2017--2025, 2015.

\bibitem[Jiang et~al.(2017)Jiang, Wu, Zheng, and Wong]{jiang2017learning}
Bai Jiang, Tung-yu Wu, Charles Zheng, and Wing~H Wong.
\newblock Learning summary statistic for approximate bayesian computation via
  deep neural network.
\newblock \emph{Statistica Sinica}, pp.\  1595--1618, 2017.

\bibitem[Khan et~al.(2016)Khan, Husa, Hannam, Ohme, P\"urrer, Jim\'nez~Forteza,
  and Boh\'e]{Khan:2015jqa}
Sebastian Khan, Sascha Husa, Mark Hannam, Frank Ohme, Michael P\"urrer, Xisco
  Jim\'nez~Forteza, and Alejandro Boh\'e.
\newblock {Frequency-domain gravitational waves from nonprecessing black-hole
  binaries. II. A phenomenological model for the advanced detector era}.
\newblock \emph{Phys. Rev.}, D93\penalty0 (4):\penalty0 044007, 2016.
\newblock \doi{10.1103/PhysRevD.93.044007}.

\bibitem[Krizhevsky et~al.(2012)Krizhevsky, Sutskever, and
  Hinton]{krizhevsky2012imagenet}
Alex Krizhevsky, Ilya Sutskever, and Geoffrey~E Hinton.
\newblock Imagenet classification with deep convolutional neural networks.
\newblock \emph{Advances in neural information processing systems},
  25:\penalty0 1097--1105, 2012.

\bibitem[Lopez{-}Paz \& Oquab(2017)Lopez{-}Paz and
  Oquab]{lopez-paz2018revisiting}
David Lopez{-}Paz and Maxime Oquab.
\newblock Revisiting classifier two-sample tests.
\newblock In \emph{5th International Conference on Learning Representations,
  {ICLR}}, 2017.

\bibitem[Lueckmann et~al.(2021)Lueckmann, Boelts, Greenberg, Goncalves, and
  Macke]{lueckmann2021benchmarking}
Jan-Matthis Lueckmann, Jan Boelts, David Greenberg, Pedro Goncalves, and Jakob
  Macke.
\newblock Benchmarking simulation-based inference.
\newblock In \emph{International Conference on Artificial Intelligence and
  Statistics}, pp.\  343--351. PMLR, 2021.

\bibitem[Papamakarios \& Murray(2016)Papamakarios and
  Murray]{papamakarios2016fast}
George Papamakarios and Iain Murray.
\newblock Fast $\varepsilon$-free inference of simulation models with bayesian
  conditional density estimation.
\newblock In \emph{Advances in neural information processing systems}, pp.\
  1028--1036, 2016.

\bibitem[Papamakarios et~al.(2019)Papamakarios, Sterratt, and
  Murray]{papamakarios2019sequential}
George Papamakarios, David Sterratt, and Iain Murray.
\newblock Sequential neural likelihood: Fast likelihood-free inference with
  autoregressive flows.
\newblock In \emph{Proceedings of the 22nd International Conference on
  Artificial Intelligence and Statistics (AISTATS)}, volume~89 of
  \emph{Proceedings of Machine Learning Research}, pp.\  837--848. PMLR, 2019.

\bibitem[Papamakarios et~al.(2021)Papamakarios, Nalisnick, Rezende, Mohamed,
  and Lakshminarayanan]{papamakarios2019normalizing}
George Papamakarios, Eric Nalisnick, Danilo~Jimenez Rezende, Shakir Mohamed,
  and Balaji Lakshminarayanan.
\newblock Normalizing flows for probabilistic modeling and inference.
\newblock \emph{Journal of Machine Learning Research}, 22\penalty0
  (57):\penalty0 1--64, 2021.
\newblock URL \url{http://jmlr.org/papers/v22/19-1028.html}.

\bibitem[Paszke et~al.(2019)Paszke, Gross, Massa, Lerer, Bradbury, Chanan,
  Killeen, Lin, Gimelshein, Antiga, Desmaison, Kopf, Yang, DeVito, Raison,
  Tejani, Chilamkurthy, Steiner, Fang, Bai, and Chintala]{pytorch}
Adam Paszke, Sam Gross, Francisco Massa, Adam Lerer, James Bradbury, Gregory
  Chanan, Trevor Killeen, Zeming Lin, Natalia Gimelshein, Luca Antiga, Alban
  Desmaison, Andreas Kopf, Edward Yang, Zachary DeVito, Martin Raison, Alykhan
  Tejani, Sasank Chilamkurthy, Benoit Steiner, Lu~Fang, Junjie Bai, and Soumith
  Chintala.
\newblock Pytorch: An imperative style, high-performance deep learning library.
\newblock In H.~Wallach, H.~Larochelle, A.~Beygelzimer, F.~d'Alch\'{e} Buc,
  E.~Fox, and R.~Garnett (eds.), \emph{Advances in Neural Information
  Processing Systems 32}, pp.\  8024--8035. Curran Associates, Inc., 2019.
\newblock URL
  \url{http://papers.neurips.cc/paper/9015-pytorch-an-imperative-style-high-performance-deep-learning-library.pdf}.

\bibitem[Rezende \& Mohamed(2015)Rezende and Mohamed]{rezende2015variational}
Danilo Rezende and Shakir Mohamed.
\newblock Variational inference with normalizing flows.
\newblock In \emph{International conference on machine learning}, pp.\
  1530--1538. PMLR, 2015.

\bibitem[Roberts \& Smith(1994)Roberts and Smith]{roberts1994simple}
Gareth~O Roberts and Adrian~FM Smith.
\newblock Simple conditions for the convergence of the gibbs sampler and
  metropolis-hastings algorithms.
\newblock \emph{Stochastic processes and their applications}, 49\penalty0
  (2):\penalty0 207--216, 1994.

\bibitem[Sisson et~al.(2018)Sisson, Fan, and Beaumont]{sisson2018handbook}
Scott~A Sisson, Yanan Fan, and Mark Beaumont.
\newblock \emph{Handbook of approximate Bayesian computation}.
\newblock CRC Press, 2018.

\bibitem[Tejero-Cantero et~al.(2020)Tejero-Cantero, Boelts, Deistler,
  Lueckmann, Durkan, Gonçalves, Greenberg, and Macke]{Tejero:2020sbi}
Alvaro Tejero-Cantero, Jan Boelts, Michael Deistler, Jan-Matthis Lueckmann,
  Conor Durkan, Pedro~J. Gonçalves, David~S. Greenberg, and Jakob~H. Macke.
\newblock sbi: A toolkit for simulation-based inference.
\newblock \emph{Journal of Open Source Software}, 5\penalty0 (52):\penalty0
  2505, 2020.
\newblock \doi{10.21105/joss.02505}.

\bibitem[Veitch et~al.(2015)]{Veitch:2014wba}
J.~Veitch et~al.
\newblock {Parameter estimation for compact binaries with ground-based
  gravitational-wave observations using the LALInference software library}.
\newblock \emph{Phys. Rev.}, D91\penalty0 (4):\penalty0 042003, 2015.
\newblock \doi{10.1103/PhysRevD.91.042003}.

\end{thebibliography}

\appendix
\counterwithin{figure}{section}
\counterwithin{table}{section}
\counterwithin{footnote}{section}

\newpage
\section{Derivations}
\subsection{Equivariance relations}
\label{sec:appendix-equivariance-relations}

Consider a system with an exact equivariance under a joint transformation of parameters $\theta$ and observations $x$,
\begin{align}
    \theta&\rightarrow \theta'=g\theta,\\
    x&\rightarrow x'=T_g x.
\end{align}
An invariant prior fulfills the relation
\begin{align}
    \label{eq:app-equiv-prior}
    p(\theta) = p(\theta') \left|\det J_g\right|,
\end{align}
where the Jacobian $J_g$ arises from the change of variables rule for probability distributions. 
A similar relation holds for an equivariant likelihood,
\begin{align}
    \label{eq:app-equiv-likelihood}
    p(x|\theta) = p(x'|\theta')\left|\det J_T\right|.
\end{align}
An invariant prior and an equivariant likelihood further imply for the evidence $p(x)$
\begin{align}
    \label{eq:app-equiv-evidence}
    p(x) = \int p(x|\theta) p(\theta) d\theta = 
    \int p(x'|\theta')\left|\det J_T\right| p(\theta') \left|\det J_g\right| \,d\theta
    = p(x') \left|\det J_T\right|.
\end{align}
Combining an invariant prior with an equivariant likelihood thus leads to the equivariance relation
\begin{align}
    \begin{split}
    p(\theta|x)
    &=\frac{p(x|\theta)p(\theta)}{p(x)}
    =\frac{p(x'|\theta')\left|\det J_T\right| p(\theta') \left|\det J_g\right|}{p(x') \left|\det J_T\right|}\\
    &=p(\theta'|x') \left|\det J_g\right|
    \end{split}
\end{align}
for the posterior, where we used Bayes' theorem and equations~\eqref{eq:app-equiv-prior},~\eqref{eq:app-equiv-likelihood}, and~\eqref{eq:app-equiv-evidence}.

\subsection{Equivariance of $p(\theta|x,\hat g)$}
\label{sec:appendix-equivariance-extended-posterior}

Here we derive that an equivariant posterior $p(\theta|x)$ remains
equivariant if the distribution is also conditioned on the proxy
$\hat g$, as used in equation~\eqref{eq:alignment}. With the
definition
$p(\hat g|\theta) = \kappa\left((g^\theta)^{-1}\hat g\right)$ from
section~\ref{subsec:gnpe}, $p(\hat g|\theta)$ is
equivariant under joint application of $h\in G$ to $\hat g$ and
$\theta$,
\begin{align}
    \label{eq:app-equiv-proxy-theta}
    p(\hat g|\theta) 
    = \kappa\left((g^\theta)^{-1}\hat g\right)
    = \kappa\left((g^\theta)^{-1}h^{-1}h\hat g\right)
    = \kappa\left((hg^\theta)^{-1}h\hat g\right)
    = p(h\hat g|h\theta).
\end{align}
where for the last equality we used 
$g^{h\theta}=h g^{\theta}$. This implies, that $p(\hat g|x)$ is equivariant under joint application of $h$ and $T_h$,
\begin{align}
    \label{eq:app-equiv-proxy-x}
    \begin{split}
    p(\hat g|x) 
    &=
    \int p(\hat g|\theta,x) p(\theta|x) \,d\theta
    \stackrel{\eqref{eq:app-equiv-proxy-theta},\eqref{eq:equiv-posterior}}{=} \int p(h\hat g|h\theta) p(h\theta|T_h x) \left|\det J_h\right| \,d\theta\\
    &= 
    p(h\hat g|T_h x).
    \end{split}
\end{align}
in the second step we used $p(\hat g|\theta,x) = p(\hat g|\theta)$. 
From these relations, the equivariance relation used in equation~\eqref{eq:alignment} follows,
\begin{align}
    \begin{split}
    p(\theta|x,\hat g) 
    &= \frac{p(\theta,\hat g|x)}{p(\hat g|x)}
    = \frac{p(\hat g|\theta,x)p(\theta|x)}{p(\hat g|x)}
    \stackrel{\eqref{eq:app-equiv-proxy-theta},\eqref{eq:equiv-posterior},\eqref{eq:app-equiv-proxy-x}}{=}
    \frac{p(h\hat g|h\theta)p(h\theta|T_h x)}{p(h\hat g|T_h x)}\left|\det J_h\right|\\
    &= p(h\theta|T_h x,h\hat g)\left|\det J_h\right|.
    \end{split}
\end{align}

\subsection{Exact equivariance of inferred posterior}
\label{sec:appendix-exact}

Consider a posterior that is exactly equivariant under $G$,
\begin{equation}\label{eq:app-equiv-posterior}
  p(\theta | x) = p(g\theta|T_gx) |\det J_g|, \qquad \forall g\in G.
\end{equation}
We here show that the posterior estimated using GNPE is equivariant under $G$ by construction. This holds regardless of whether $q(\theta'|x')$ has fully converged to $p(\theta'|x')$.

With GNPE, the equivariant posterior $p(\theta|x_o)$ for an observation $x_o$ is inferred by alternately sampling 
\begin{align}
  \label{eq:app-Gibbs-theta}
  \theta^{(i)} &\sim p(\theta|x_o, \hat g^{(i-1)}) & &\Longleftrightarrow& \theta^{(i)} &= \hat g^{(i-1)} \theta'^{(i)}, \quad \theta'^{(i)} \sim q(\theta' | T_{(\hat g^{(i-1)})^{-1}} x_o),\\
  \label{eq:app-Gibbs-proxy}
  \hat g^{(i)} &\sim p(\hat g|x_o, \theta^{(i)}) & &\Longleftrightarrow& \hat g^{(i)} &= g^{\theta^{(i)}} \epsilon, \quad \epsilon \sim \kappa(\epsilon),
\end{align} 
see also equation~\eqref{eq:gnpe-exact-equiv}.
Now consider a different observation $\tilde x_o = T_h x_o$ that is obtained by altering the pose of $x_o$ with $T_h$, where $h$ is an arbitrary element of the equivariance group $G$. 
Applying the joint transformation 
\begin{align}
    \label{eq:app-proxy-equiv-trafo-1}
    \hat g&\rightarrow h\hat g,\\
    \label{eq:app-obs-equiv-trafo-1}
    x_o&\rightarrow T_h x_o,
\end{align}
in~\eqref{eq:app-Gibbs-theta} leaves $\theta'$ invariant,
\begin{align}
    q(\theta' | T_{(h\hat g)^{-1}} T_h x_o)
    = q(\theta' | T_{\hat g^{-1}} (T_h)^{-1} T_h x_o)
    = q(\theta' | T_{\hat g^{-1}} x_o).
\end{align}
We thus find that $\theta$ in~\eqref{eq:app-Gibbs-theta} transforms equivariantly under joint application of~\eqref{eq:app-proxy-equiv-trafo-1} and~\eqref{eq:app-obs-equiv-trafo-1},
\begin{align}
    \theta = \hat g\theta'\rightarrow (h\hat g)\theta'=h(\hat g\theta')=h\theta.
\end{align}
Conversely, applying
\begin{align}
    \theta\rightarrow h\theta
\end{align}
in~\eqref{eq:app-Gibbs-proxy} transforms $\hat g$ by
\begin{align}
    \hat g = g^\theta\epsilon \rightarrow g^{(h\theta)}\epsilon = h g^\theta\epsilon = h \hat g.
\end{align}
The $\theta$ samples (obtained by marginalizing over $\hat g$) thus transform $\theta\rightarrow h\theta$ under $x_o\rightarrow T_h x_0$, which is consistent with the desired equivariance~\eqref{eq:app-equiv-posterior}.

Another intuitive way to see this is to consider running  an implementation of the Gibbs sampling steps~\eqref{eq:app-Gibbs-theta} and~\eqref{eq:app-Gibbs-proxy} with fixed random seed for two observations $x_o$ (initialized with $\hat g^{(0)}=\hat g_{x_o}$) and $T_h x_o$ (initialized with $\hat g^{(0)}=h\hat g_{x_o}$). The Gibbs sampler will yield parameters samples $(\theta_i)_{i=1}^N$ for $x_o$, and the \emph{exact same} samples $(h\theta_i)_{i=1}^N$ for $T_h x_o$, up to the global transformation by $h$. The reason is that the density estimator $q(\theta'|x')$ is queried with the same $x'$ for both observations $x_o$ and $T_h x_o$ in each iteration $i$. Since the truncated, thinned samples are asymptotically independent of the initialization, this shows that~\eqref{eq:app-equiv-posterior} is fulfilled by construction.

\subsection{Iterative inference and convergence}
\label{sec:appendix-iterative-inference}
GNPE leverages a neural density estimator of the form $q(\theta|x',\hat g)$ to obtain samples from the joint distribution $p(\theta,\hat g|x)$. This is done by iterative sampling as described in section~\ref{sec:methods}. Here we derive equation~\eqref{eq:convergence}, which states how a distribution $Q_{j}(\theta|x)$ is updated by a single GNPE iteration. 

Given a distribution $Q_{j}^\theta(\theta|x)$ of $\theta$ samples in iteration $j$, we infer samples for the pose proxy $\hat g$ for the next iteration by (i) extracting the pose $g^\theta$ from $\theta$ (this essentially involves marginalizing over all non pose related parameters) and (ii) blurring the pose $g^\theta$ with the kernel $\kappa$, corresponding to a group convolution.\footnote{We define a group convolution as $(A\ast B)(\hat g)=\int dg\, A(g) B(g^{-1}\hat g)$, which is the natural extension of a standard convolution.}
We denote this combination of marginalization and group convolution with the ``$\cm$'' symbol,
\begin{equation}
    \label{eq:proxy-distribution-from-theta-distribution}
    Q_{j+1}^{\hat g}(\hat g|x)
    = \int d\theta\, Q^\theta_{j}(\theta|x) \kappa((g^\theta)^{-1}\hat g) 
    = \left(Q^\theta_{j}(\cdot|x)\cm \kappa\right)(\hat g).
\end{equation}
For a given proxy sample $\hat g$, a (perfectly trained) neural density estimator infers $\theta$ with
\begin{align}
    \label{eq:theta-posterior-given-proxy}
    p(\theta|x,\hat g) 
    &= \frac{p(\theta,\hat g|x)}{p(\hat g|x)}
    = \frac{p(\hat g|x,\theta) p(\theta|x)}{p(\hat g|x)}
    = p(\theta|x)\frac{\kappa((g^\theta)^{-1} \hat g)}{\left(p^\theta(\cdot|x)\cm \kappa\right)(\hat g) },
\end{align}
where we used $p(\hat g| \theta) = \kappa((g^\theta)^{-1} \hat g)$. Combining~\eqref{eq:proxy-distribution-from-theta-distribution} and~\eqref{eq:theta-posterior-given-proxy}, the updated distribution over $\theta$ samples reads
\begin{equation}
    \begin{split}
        Q_{j+1}^\theta(\theta|x)&=\int d\hat g\, p(\theta|x,\hat g)\, Q_{j+1}^{\hat g}(\hat g|x)\\
        &= \int d\hat g\, 
        \left(Q^\theta_{j}(\cdot|x)\cm \kappa\right)(\hat g)
        \,p(\theta|x)\frac{\kappa((g^\theta)^{-1} \hat g)}{\left((p^\theta(\cdot|x)\cm \kappa\right))(\hat g) }\\
        &= p(\theta|x)\,\int d\hat g\, 
        \frac{\left(Q^\theta_{j}(\cdot|x)\cm \kappa\right)(\hat g)}{\left((p^\theta(\cdot|x)\cm \kappa\right))(\hat g) }
        \,\kappa((g^\theta)^{-1} \hat g)\\
        &= p(\theta|x)\, 
        \left(
        \frac{Q^\theta_{j}(\cdot|x)\cm \kappa}{p^\theta(\cdot|x)\cm \kappa}
        \,\ast\,
        \,\kappa^{(-)}\right)(\hat g).
    \end{split}
\end{equation}
Here, $\kappa^{(-)}$ denotes the reflected kernel, $\kappa^{(-)}(g)=\kappa(g^{-1})\,\forall g$. Since we choose a symmetric kernel in practice, we use $\kappa=\kappa^{(-)}$ in~\eqref{eq:convergence}.

In this notation, the initialization of the pose $g^\theta$ in iteration $0$ with $q_\text{init}$ simply means setting $Q_0(\cdot|x) \cm \kappa = q_\text{init}(\cdot|x) \ast \kappa$.

\section{GNPE for simple Gaussian likelihood and prior}
\label{sec:app-Gaussian-likelihood-prior}
Consider a simple forward model $\tau\rightarrow x$ with a given prior
\begin{align}\label{eq:def-prior}
    p(\tau) = \N (-5,1)[\tau]
\end{align}
and a likelihood
\begin{align}\label{eq:def-likelihood}
    p(x|\tau) = \N(\tau,1)[x],
\end{align}
where the normal distribution is defined by
\begin{align}\label{eq:def-normal}
    \N(\mu,\sigma^2)[x] = \frac{\exp\left(\frac{-(x-\mu)^2}{2\sigma^2}\right)}{\sqrt{2\pi}\sigma}.
\end{align}
The evidence can be computed from the prior~\eqref{eq:def-prior} and likelihood~\eqref{eq:def-likelihood}, and reads
\begin{align}\label{eq:evidence}
    p(x) 
    = \int d\tau p(\tau)p(x|\tau) 
    = \int d\tau \N (-5,1)[\tau] \N(\tau,1)[x]
    = \N(-5,2)[x].
\end{align}
The posterior is then given via Bayes' theorem and reads
\begin{align}\label{eq:posterior}
    p(\tau|x) = \frac{p(x|\tau)p(\tau)}{p(x)}
    = \frac{\N(\tau,1)[x]\N (-5,1)[\tau]}{\N(-5,\sqrt{2})[x]}
    = \N\left(\frac{x-5}{2},1/2\right)[\tau].
\end{align}
\subsection{Equivariances}
The likelihood~\eqref{eq:def-likelihood} is equivariant under $G$, i.e., the joint transformation
\begin{align}
    \begin{split}
        \tau&\rightarrow g\tau = \tau + \Delta\tau,\\
        x&\rightarrow T^{l}_g x = x + \Delta\tau.
    \end{split}
\end{align}
This follows directly from~\eqref{eq:def-likelihood} and~\eqref{eq:def-normal}. 
If the prior was invariant, then this equivariance would be inherited by the posterior, see App.~\ref{sec:appendix-equivariance-relations}. However, the prior is not invariant. It turns out that the posterior is still equivariant, but $x$ transforms under a different representation than it does for the equivariance of the likelihood. Specifically, the posterior~\eqref{eq:posterior} is equivariant under joint transformation 
\begin{align}\label{eq:joint-transformation-posterior}
    \begin{split}
        \tau&\rightarrow g\tau = \tau + \Delta\tau,\\
        x&\rightarrow T^{p}_g x = x + 2\cdot\Delta\tau,
    \end{split}
\end{align}
which again directly follows from~\eqref{eq:posterior} and~\eqref{eq:def-normal}. 
Importantly, $T_g^p\neq T_g^l$, i.e., the representation under which $x$ transforms is different for the equivariance of the likelihood and the posterior. For GNPE, the relevant equivariance is that of the posterior, i.e. the set of transformations~\eqref{eq:joint-transformation-posterior}, see also equation~\eqref{eq:equiv-posterior}. 
The equivariance relation of the posterior thus reads
\begin{align}\label{eq:equivariance-posterior}
     p(\tau | x) = p(g\tau|T^p_gx) |\det J_g|, \qquad \forall g\in G.
\end{align}

\subsection{GNPE}
\begin{figure}
    \centering
    \includegraphics[width=0.8\textwidth]{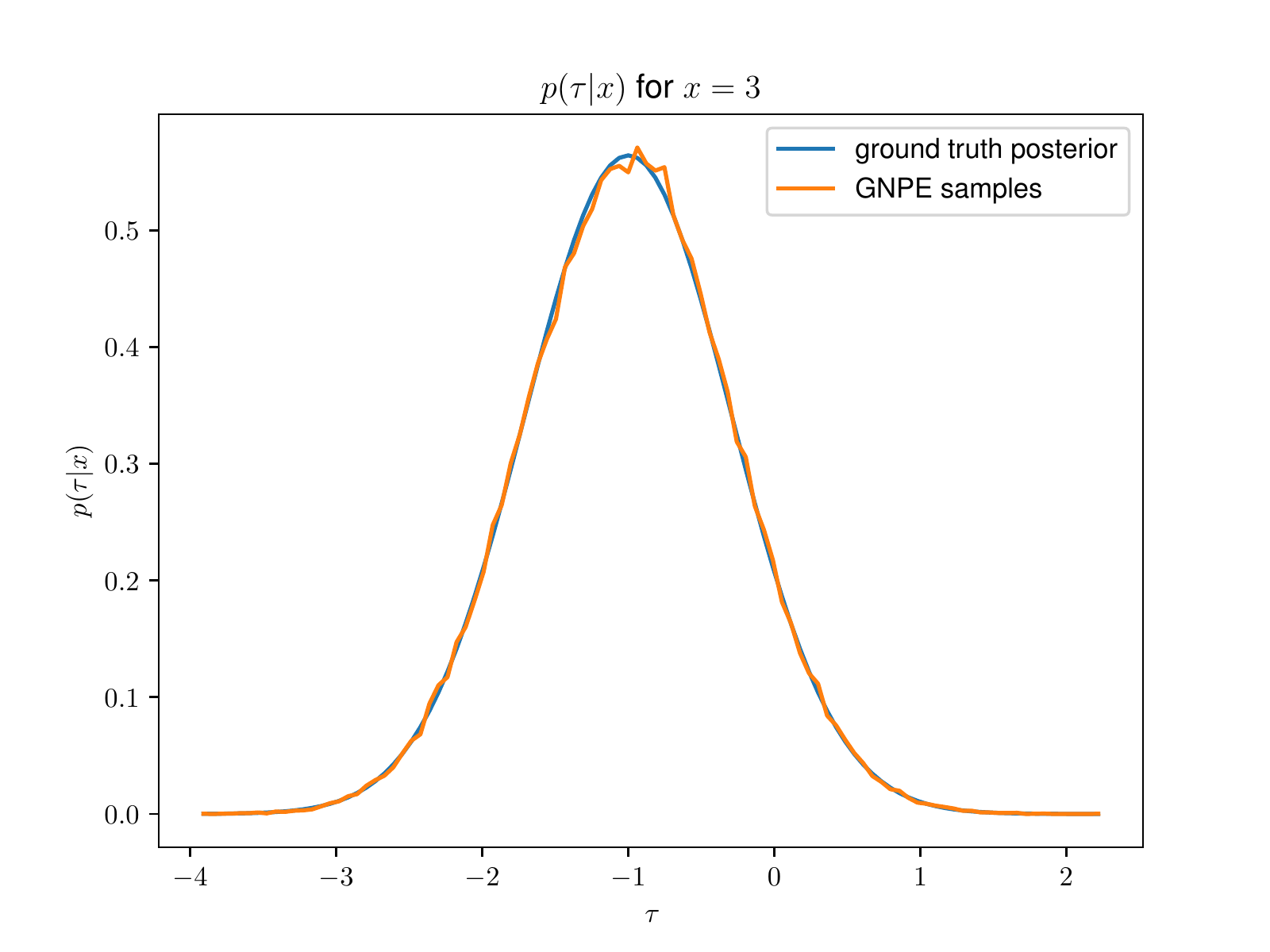}
    \caption{Posterior $p(\tau|x=3)$ (blue) and the corresponding inferred GNPE samples (orange).}
    \label{fig:simple-toy-example-result}
\end{figure}

We choose $\tau$ as the pose, which we aim to standardize with GNPE. We define the corresponding proxy as
\begin{align}
    \hat\tau = \tau + \epsilon,\quad \epsilon\sim \kappa(\epsilon)=\N(0,1)[\epsilon].
\end{align}
We can use GNPE to incorporate the exact equivariance of the posterior by construction. To that end we define
\begin{align}\label{eq:tauprime-xprime}
    \begin{split}
    \tau' &= g^{(-\hat\tau)}\tau =  \tau -\hat\tau,\\
    x' &= T^p_{g^{(-\hat\tau)}}x = x - 2\cdot\hat \tau.
    \end{split}
\end{align}
We then train a neural density estimator to estimate $p(\tau'|x')$. This distribution is of the same form as $p(\tau|x)$ and simply given by
\begin{align}\label{eq:posterior-tauprime-given-xprime}
    p(\tau'|x')
    \stackrel{\eqref{eq:equivariance-posterior},\eqref{eq:posterior}}{=} 
    \N\left(\frac{x'-5}{2}, 1/2 \right)[\tau']
\end{align}
due to the equivariance~\eqref{eq:equivariance-posterior}. We here assume a neural density estimator that estimates~\eqref{eq:posterior-tauprime-given-xprime} perfectly. 
For GNPE, we
\begin{enumerate}
    \item Initialize $\tau^{(1)}=0$;
    \item Sample $\hat\tau^{(1)}$ by $\hat\tau^{(1)}=\tau^{(1)}+\epsilon,~\epsilon\sim\N(0,1)[\epsilon]$, and compute $\tau'$ and $x'$ via~\eqref{eq:tauprime-xprime};
    \item Sample $\tau^{(2)}$ by $\tau^{(2)}=\tau'^{(2)}+\hat\tau^{(1)}$, with $\tau'^{(2)}\sim p(\tau'|x')=\N\left(\frac{x'-5}{2}, 1/2 \right)[\tau']$;
\end{enumerate}
and repeat (2) and (3) multiple times. This constructs a Markov chain. To obtain (approximately independent) posterior samples $\tau\sim p(\tau|x)$, we truncate to account for burn-in, thin the chain and marginalize over $\hat\tau$. We find that the chain indeed converges to the correct posterior~\eqref{eq:posterior}, see Fig.~\ref{fig:simple-toy-example-result}.

\section{Toy Example}
\label{sec:appendix-toy-example}

\subsection{Forward model}
\label{sec:appendix-toy-example-forward-model}
The toy model in section~\ref{sec:toy-example} describes the motion of a damped harmonic oscillator that is initially at rest and excited at time $\tau$ with an infinitely short pulse. The time evolution of that system is governed by the differential equation
\begin{align}
    \label{eq:ho-diffeq}
    \frac{d^2}{dt^2}x(t)+2\beta \omega_0\frac{d}{dt}x(t)+\omega_0^{2}x(t)=\delta(t-\tau),
\end{align}
where $\omega_0$ denotes the undamped angular frequency and $\beta$ the damping ratio. The solution for the time series $x(t)$ is given by the Green's function for the corresponding differential operator and reads
\begin{align}
    \label{eq:ho-solution}
    x(t)=
    \begin{cases}
        0, &t\leq\tau\\
        e^{-\beta\omega_0(t-\tau)}\cdot\frac{\sin\left(\sqrt{1-\beta^2}\omega_0 (t-\tau)\right)}{\sqrt{1-\beta^2}\omega_0}, &t>\tau.
    \end{cases}
\end{align}
This equation describes a deterministic, injective mapping between parameters $\theta=(\omega_0,\beta,\tau)$ and a time series observation $x$,
\begin{equation}
    x = f(\theta).
\end{equation}
This implies a likelihood $p(x|\theta)=\delta(x-f(\theta))$, and thus a point-like posterior. To showcase (G)NPE on this toy problem we introduce stochasticity by setting 
\begin{equation}
    x = f(\theta + \delta\theta)
\end{equation}
instead. We sample $\delta\theta$ from an uncorrelated Gaussian distribution
\begin{equation}
    \delta\theta\sim \mathcal{N}(0, \Sigma),
    \qquad \Sigma = 
    \left(\begin{array}{ccc}
        \sigma_{\omega_0}^2 & 0 & 0 \\
        0 & \sigma_{\beta}^2 & 0 \\
        0 & 0 & \sigma_{\tau}^2 \\
    \end{array}\right),
\end{equation}
with $\sigma_{\omega_0}=0.3$~Hz, $\sigma_\beta=0.03$ and $\sigma_\tau=0.3$~s. Due to the injectivity of $f$, the posterior $p(\theta|x)$ reduces to the probability $p(\delta\theta=f^{-1}(x)-\theta)$. With a uniform prior, and neglecting boundary effects, this implies an uncorrelated Gaussian posterior $p(\theta|x)$ centered around $f^{-1}(x)$ with standard deviations as specified above. We choose this approach over, e.g., adding noise straight to observations to keep the problem as simple as possible, such that the focus remains on the comparison of GNPE and NPE. In particular, knowing that the ground truth posteriors are Gaussian, we can use a simple Gaussian density estimator.

We choose uniform priors
\begin{equation}
    p(\omega_0)=\mathcal{U}[3,10]~\text{Hz},
    \qquad p(\beta)=\mathcal{U}[0.2,0.5],
    \qquad p(\tau)=\mathcal{U}[-5,0]~\text{s}.
\end{equation}
The observational data $x$ is the discretized time series in the interval $[-5,+5]$~s with 2000 evenly sampled bins. When applying time shifts with GNPE, we impose cyclic boundary conditions.

\subsection{Implementation}
\label{sec:appendix-toy-example-implementation}
We use a Gaussian density estimator for all methods (since we know that the true posterior is Gaussian). For NPE, we use a feedforward neural network with [128, 32, 16] hidden units and with ReLU activation functions as an embedding network. For NPE-CNN, we use a three-layer convolutional embedding network with kernel sizes [5,5,5], stride 1, [6,12,12] channels, average pooling with kernel size 7 and stride 7, and ReLU activation functions. For GNPE, we use the same architecture as for NPE for both, $q(\theta'|x')$ and $q_\text{init}(\tau|x)$. For further hyperparameters, we use the defaults of the sbi package~\citep{Tejero:2020sbi}.

\subsection{Results}
For all methods, we compute the average classifier two-sample test score (c2st) based on 10,000 samples from the estimated and the ground truth posterior for five different simulations. We then average the accuracy across 10 different seeds.

\section{Gravitational wave parameter inference}
\label{sec:appendix-gwpe}

\subsection{Forward model and amortization}
The forward model mapping binary black hole parameters $\theta$ (Tab.~\ref{tab:GW-parameters-with-priors}) to simulated measurements $x$ in the detectors consists of two stages. Firstly, the waveform polarizations $h(\theta)$ for given parameters $\theta$ are computed with the waveform model IMRPhenomPv2~\citep{Hannam:2013oca,Khan:2015jqa,Bohe:2016}. Secondly, the signals are projected onto the detectors, and noise is added to obtain a realistic signal $x$. To a good approximation, we assume the noise to be Gaussian and stationary over the duration of a single GW signal. However, the noise spectrum, determined by the power spectral density (PSD) $S_\text{n}$, drifts over the duration of an observing run. To fully amortize the computational cost, we use a variety of different PSDs $S_\text{n}$ in training, and additionally condition the inference network on $S_\text{n}$. At inference time, this enables instant tuning of the inference network to the PSD estimated at the time of the event, see~\citet{dax2021real} for details. Since this conditioning on $S_\text{n}$ has no effect on the GNPE algorithm outlined in this work, we keep it implicit in all equations.

\begin{table}
    \centering
    \caption{
    Priors for the astrophysical binary black hole parameters used to train the inference network. Priors are uniform over the specified range unless indicated otherwise. We train networks with different distance ranges for the two observing runs O1 and O2 due to the different detector sensitivities. 
    At inference time, a cosmological distance prior is imposed by reweighting samples according to their distance. 
    }
    \begin{tabular}{lll}
        \hline\hline
        Description & Parameter & Prior\\\hline
        component masses & $m_1$, $m_2$ & $[10,80]~\mathrm{M}_\odot$, $m_1\geq m_2$ \\
        spin magnitudes & $a_1$, $a_2$ & $[0,0.88]$\\ 
        spin angles & $\theta_1$, $\theta_2$, $\phi_{12}$, $\phi_{JL}$ & standard as in~\citet{Farr:2014qka}\\
        time of coalescence & $t_c$ & $[-0.1,0.1]$~s\\
        luminosity distance & $d_L$ & 
        \makecell[lt]{
        O1, 2 detectors: $[100,2000]$~Mpc\\
        O2, 2 detectors: $[100,2000]$~Mpc and $[100,6000]$~Mpc\\
        O2, 3 detectors: $[100,1000]$~Mpc
        }\\
        reference phase & $\phi_c$ & $[0,2\pi]$\\
        inclination & $\theta_{JN}$ & $[0,\pi]$ uniform in sine \\
        polarization & $\psi$ & $[0,\pi]$ \\
        sky position & $\alpha, \beta$ & uniform over sky
        \\\hline\hline
    \end{tabular}
    \label{tab:GW-parameters-with-priors}
\end{table}

\subsection{Network architecture and training}
\label{sec:appendix-GW-network-architecture}
The inference network consists of an embedding network, that reduces the high dimensional input data to a 128 dimensional feature vector, and the normalizing flow, that takes this feature vector as input. For each detector, the input to the embedding network consists of the complex-valued frequency domain strain in the range $[20~\text{Hz},1024~\text{Hz}]$ with a resolution of $0.125$~Hz, and PSD information $(10^{46}\cdot S_\text{n})^{-1/2}$ with the same binning. This results to a total of (3$\,\cdot\,$8,033)\,=\,24,099 real input bins per detector. The first module of the embedding network consists of a linear layer per detector, that maps this 24,096 dimensional input to 400 components. We initialize this compression layer with PCA components of raw waveforms. This provides a strong inductive bias to the network to filter out GW signals from extremely noisy data. Note that this important step is only possible since GNPE is architecture independent---it is for instance not compatible with a convolutional neural network. Following this compression layer, we use a series of 24 fully-connected residual blocks with two layers each to compress the output to the desired 128 dimensional feature vector. We use batch normalization and ELU activation functions. Importantly, the conditioning of the flow on the proxy $\hat g_\text{rel.}$ is done \emph{after} the embedding network, by concatenating $\hat g_\text{rel.}$ to the embedded feature vector. 

Following this, we use a neural spline flow~\citep{durkan2019neural} with rational-quadratic spline coupling transforms as density estimator. We use 30 such transforms, each of which is associated with 5 two-layer residual blocks with hidden dimension 512. In total, the inference network has 348 hidden layers and $1.31\cdot 10^8$ (for two detectors) or $1.42\cdot 10^8$ (for three detectors) learnable parameters.

We train the inference network with a data set of $5\cdot 10^6$ waveforms with parameters $\theta$ sampled from the priors specified in table~\ref{tab:GW-parameters-with-priors}, and reserve 2\% of the data for validation. We pretrain the network with learning rate of $3\cdot 10^{-4}$ for 300 epochs with fixed PSD, and finetune for another 150 epochs with learning rate of $3\cdot 10^{-5}$ with varying PSDs. With batch size 4,096, training takes 16-18 days on a NVIDIA Tesla V100 GPU.

\subsection{Results}
\begin{figure}
    \centering
    \includegraphics[trim={0.0cm 0.0cm 0.0cm 0.0cm},clip,width=\textwidth]{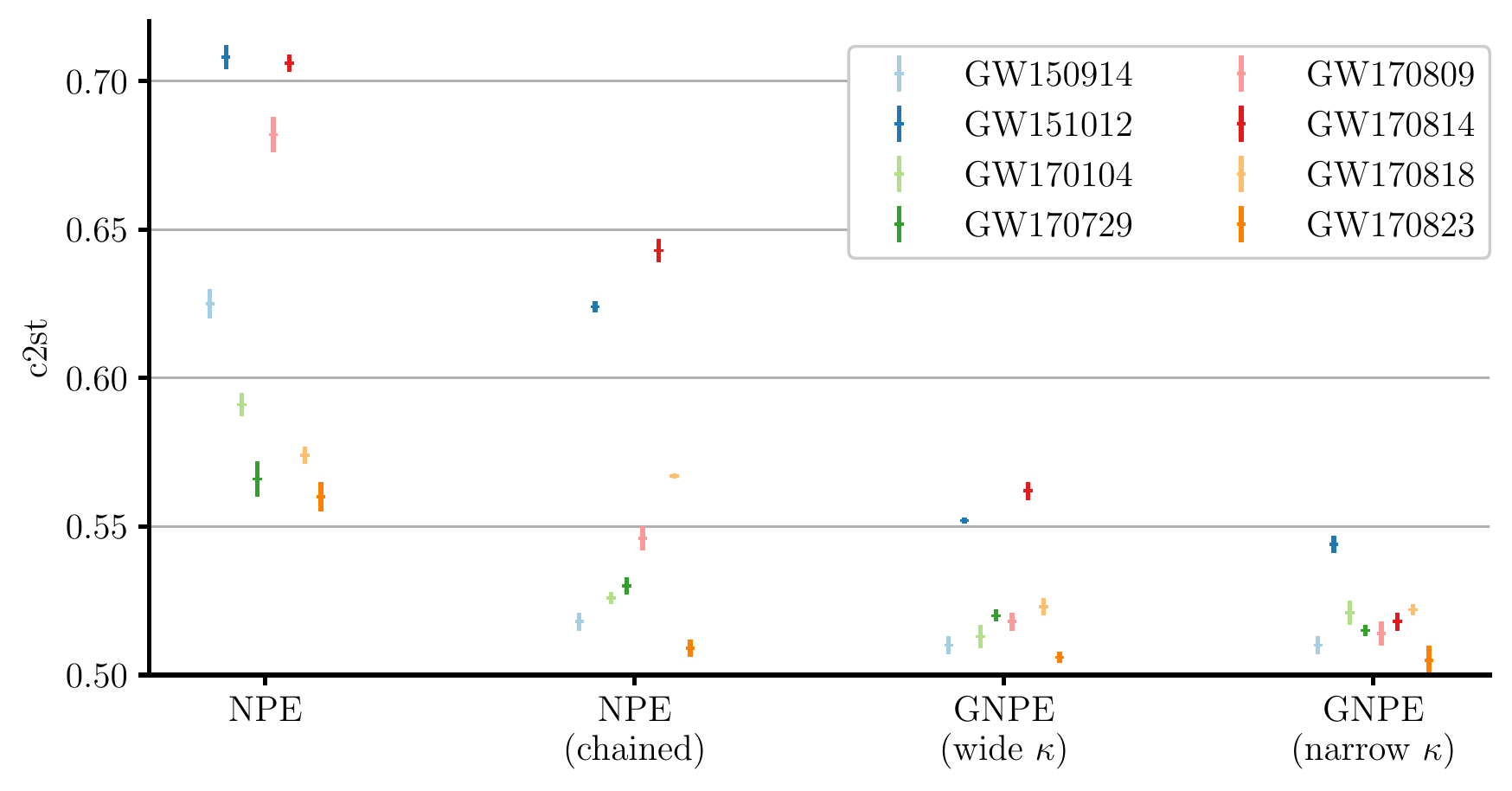}
    \caption{c2st scores quantifying the deviation between the inferred posteriors and the MCMC reference. This is an extended version of Fig.~\ref{fig:GW-ablation-studies-c2st}.}
    \label{fig:GW-ablation-studies-c2st-errorbars}
\end{figure}

The c2st scores between inferred posterior and the MCMC reference shown in Fig.~\ref{fig:GW-ablation-studies-c2st} are computed using the code and default hyperparameters of~\citet{lueckmann2021benchmarking}. For each event, we compute the c2st score of 10,000 samples for inferred and target posterior. Fig.~\ref{fig:GW-ablation-studies-c2st} displays the mean of the score across 5 different sample realizations, Fig.~\ref{fig:GW-ablation-studies-c2st-errorbars} additionally shows the corresponding standard deviation. For technical reasons we use only 12 of the 15 inferred parameters; specifically we omit the geocentric time of coalescence $t_c$ (since the reference posteriors generated with LALInference do not contain that variable) and the sky position parameters $\alpha$ and $\delta$ (since the NPE baseline with chain rule decomposition infers these in another basis). 

\begin{figure}
    \centering
    \begin{minipage}[c]{0.60\textwidth}
    \includegraphics[trim={0cm 0cm 0cm 0cm},clip,width=\textwidth]{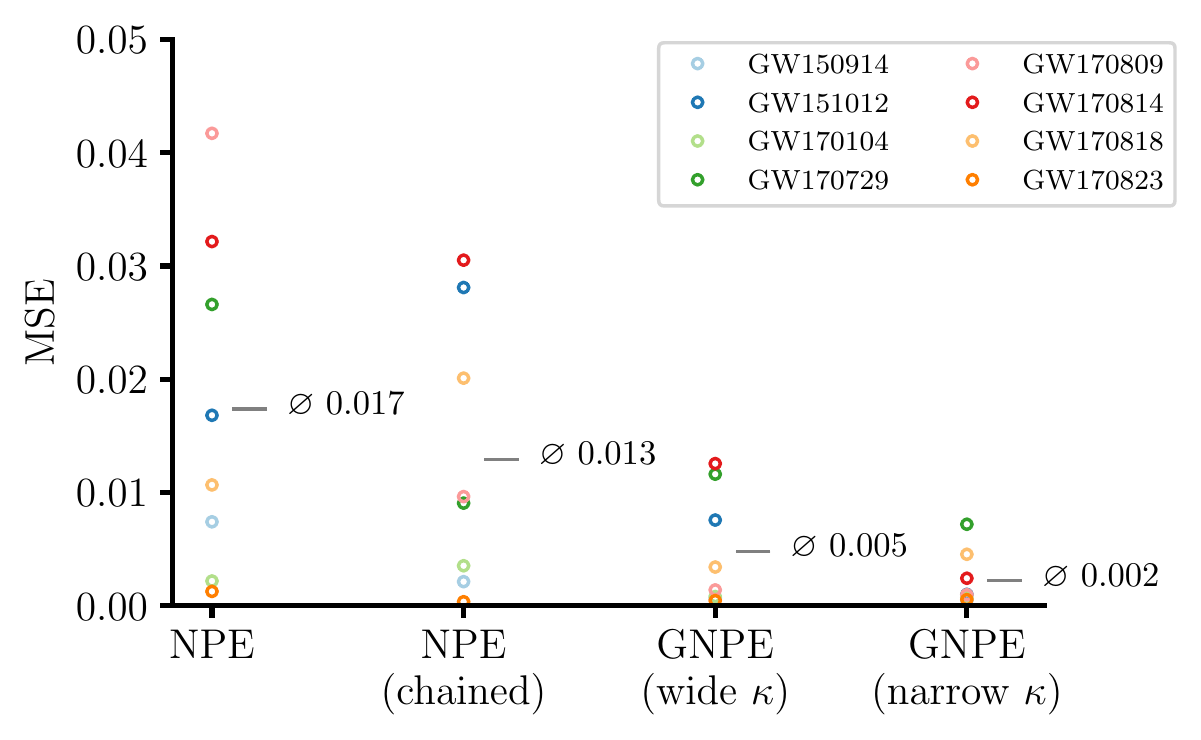}
    \end{minipage}\hfill
    \begin{minipage}[c]{0.35\textwidth}
    \caption{Comparison of estimated posteriors against \textsc{LALInference} MCMC for eight GW events, as quantified by the mean squared error (MSE) of the sample means. Before computing the means, we normalize each dimension such that the prior has a standard deviation of $1$. $\varnothing$ indicates the average across all eight events. GNPE with a narrow kernel consistently outperforms the baselines, which is in accordance with Fig.~\ref{fig:GW-ablation-studies-c2st}.
    }
    \label{fig:GW-ablation-studies-mse}
    \end{minipage}
\end{figure}

\begin{figure}
    \centering
    \includegraphics[trim={0.0cm 0.0cm 0.0cm 0.0cm},clip,width=\textwidth]{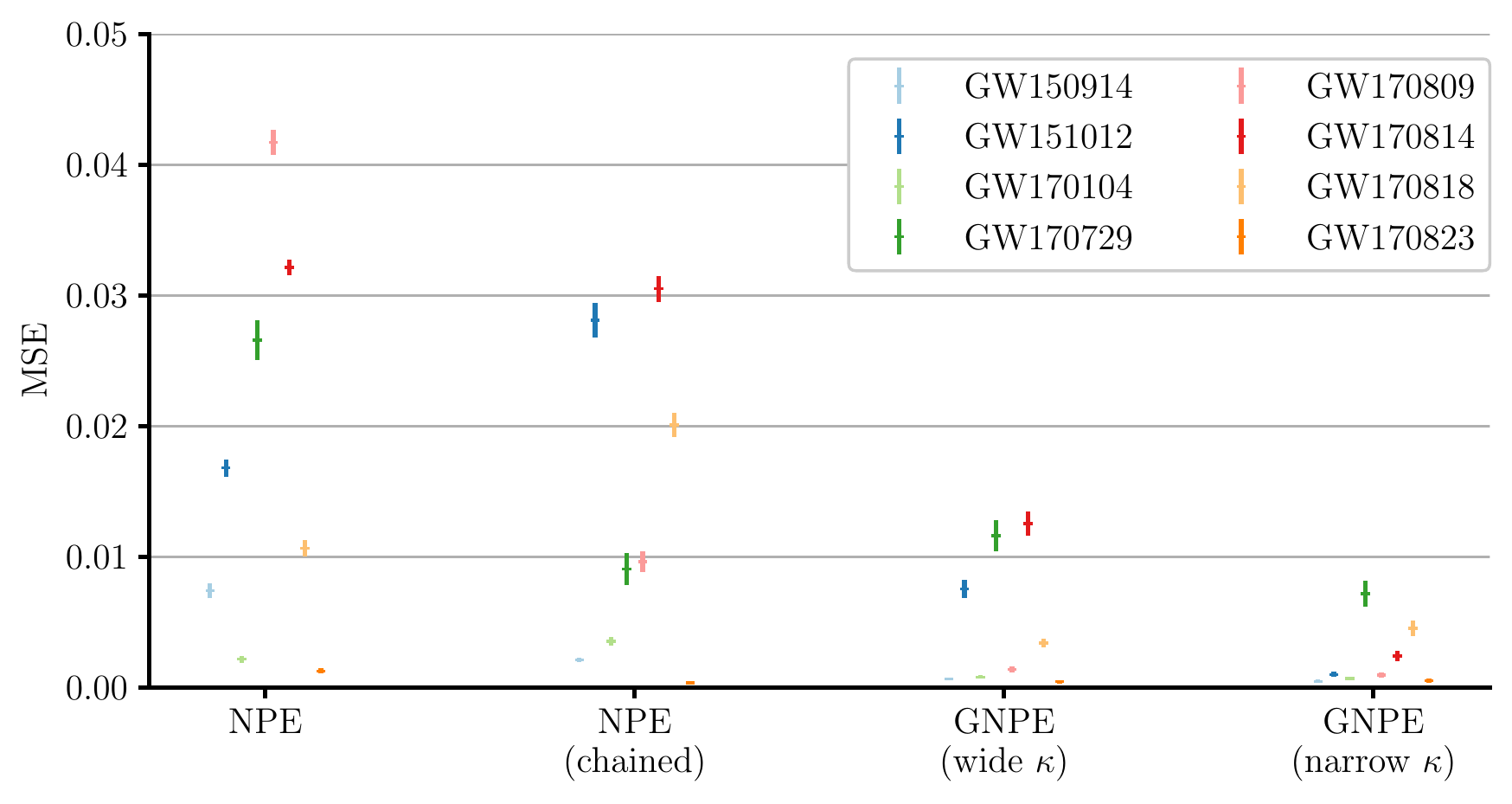}
    \caption{MSE between inferred posteriors and MCMC reference. Extended version of Fig.~\ref{fig:GW-ablation-studies-mse}.
    }
    \label{fig:GW-ablation-studies-mse-errorbars}
\end{figure}

\begin{figure}
    \centering
    \begin{subfigure}{0.48\textwidth}
        \includegraphics[width=\textwidth]{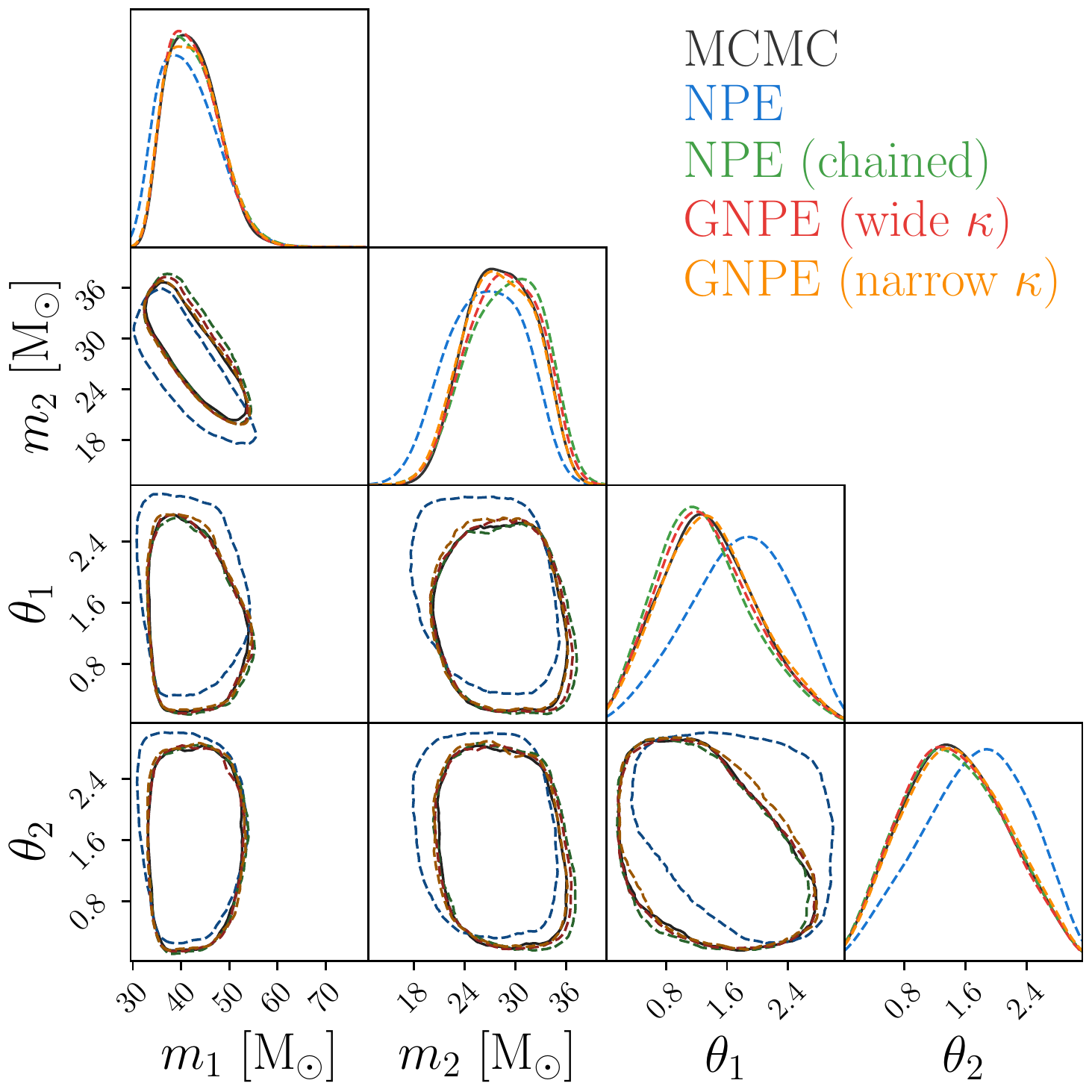}
    \end{subfigure}
    \hfill
    \begin{subfigure}{0.48\textwidth}
        \includegraphics[width=\textwidth]{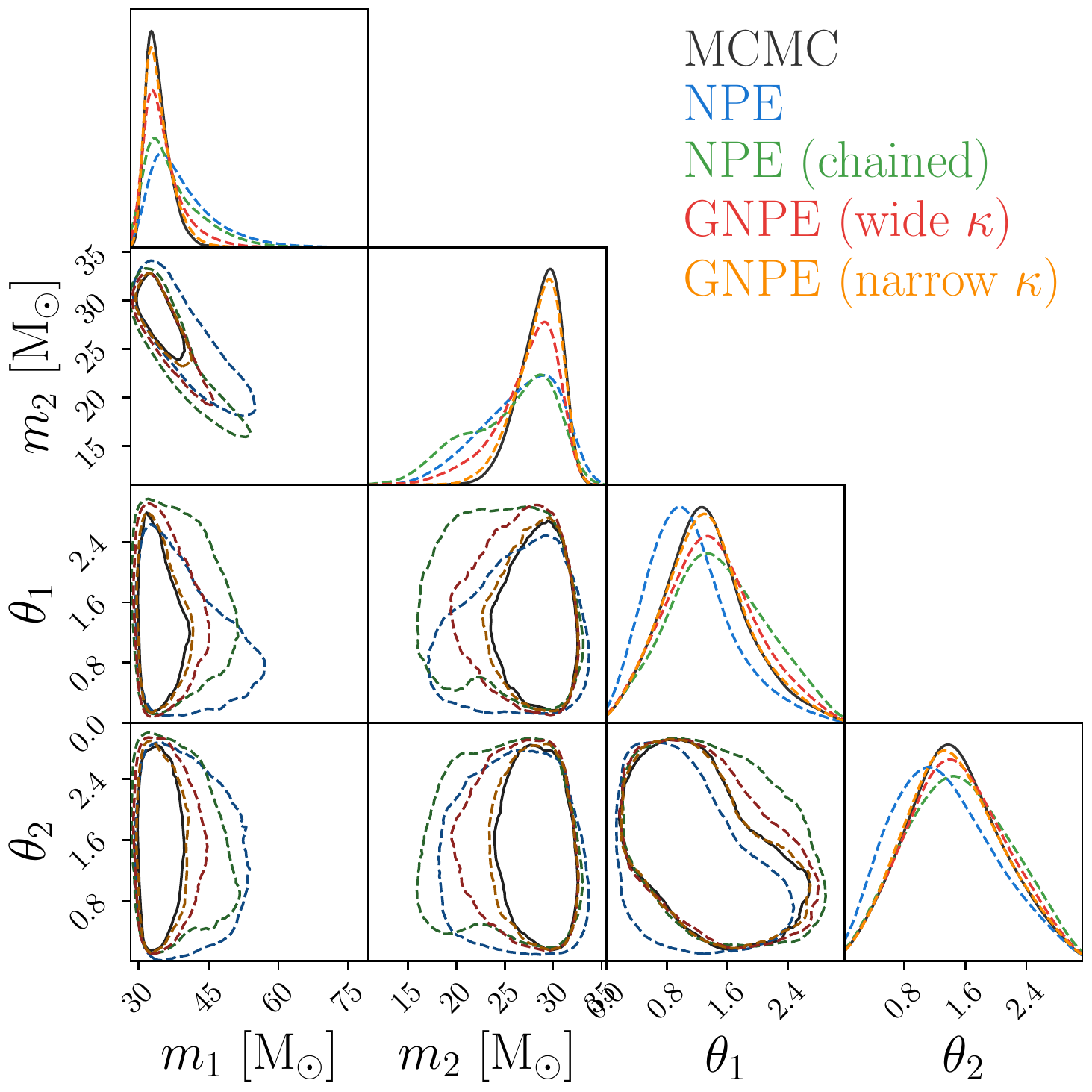}
    \end{subfigure}
    \caption{Corner plots for the GW events GW170809 (left) and
      GW170814 (right), plotting 1D marginals on the diagonal and 90\%
      credible regions for the 2D correlations. We display the two
      black hole masses $m_1$ and $m_2$ and two spin parameters
      $\theta_1$ and $\theta_2$ (note that the full posterior is
      15-dimensional). This extends Fig.~\ref{fig:GW-corner-ablation-studies} by also displaying the results from chained NPE and GNPE with wide $\kappa$. 
      }
    \label{fig:GW-corner-ablation-studies_all}
\end{figure}

\begin{figure}
    \centering
    \includegraphics[width=\textwidth]{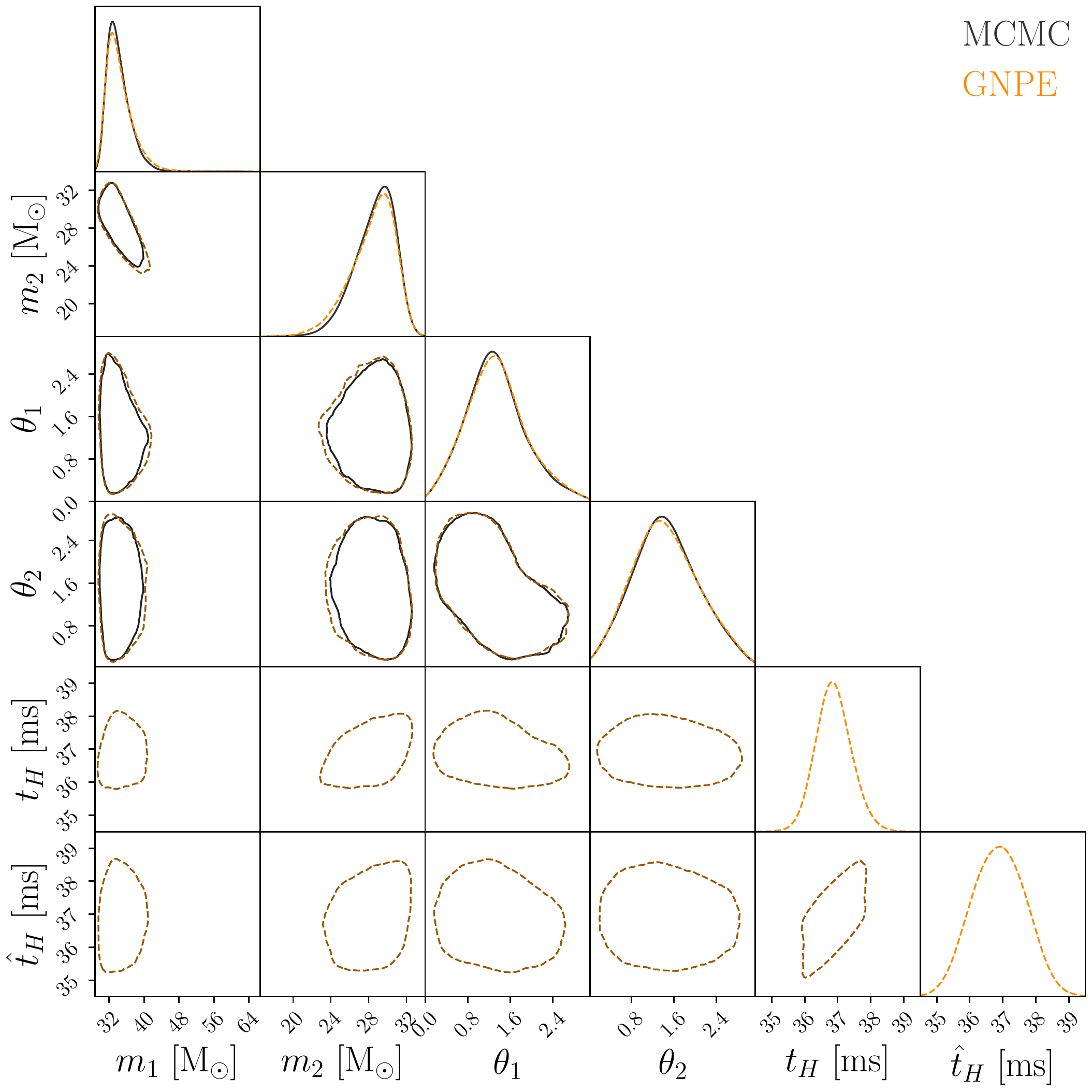}
    \caption{Corner plot for the GW event GW170814 plotting 1D marginals on the diagonal and 90\%
      credible regions for the 2D correlations. We display the two
      black hole masses $m_1$ and $m_2$ and two spin parameters
      $\theta_1$ and $\theta_2$ that are also shown in Fig.~\ref{fig:GW-corner-ablation-studies}.
      We additionally display one of the pose parameters $t_H$ and the corresponding proxy $\hat t_H$ from the last GNPE iteration. 
      In training, the neural density estimator learned that the true pose $t_H$ differs by at most $1$~ms from the proxy $\hat t_H$ that it is conditioned on (since we chose a kernel $\kappa_\text{narrow} = U[-1~\mathrm{ms},1~\mathrm{ms}]^{n_I}$, see section~\ref{sec:gwpe-gnpe}). This explains the strong correlation between $t_H$ and $\hat t_H$ we observe. For the same reason, the observed correlations between the $\hat t_H$ and the non-pose parameters $(m_1,m_2,\theta_1,\theta_2)$ are similar to those between the true pose $t_H$ and $(m_1,m_2,\theta_1,\theta_2)$. 
    }
    \label{fig:GW-corner-GNPE-pose-and-proxy}
\end{figure}

\begin{figure}
    \centering
    \includegraphics[trim={0.0cm 0.0cm 0.0cm 0.0cm},clip,width=\textwidth]{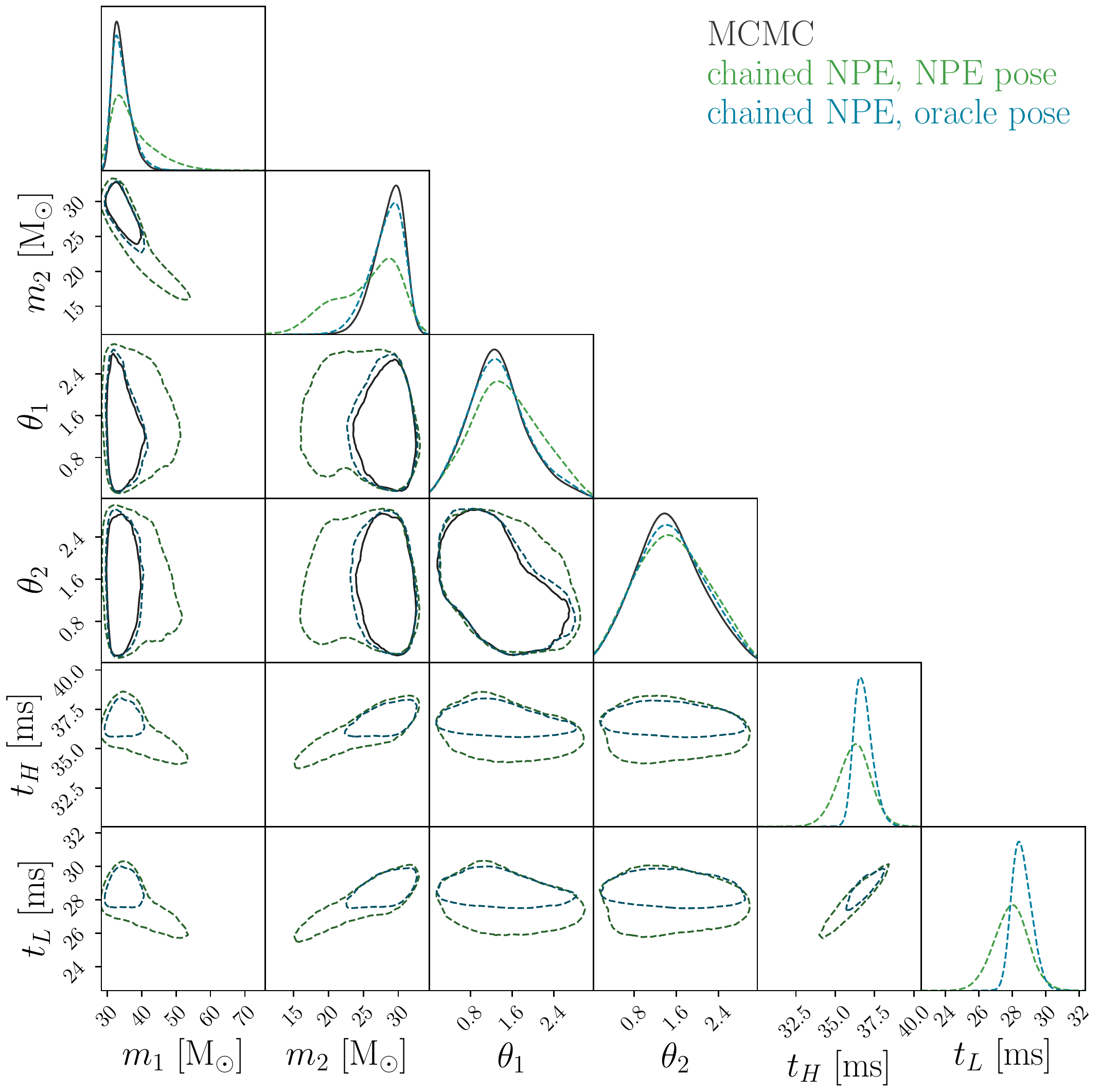}
    \caption{Corner plot for GW170814 with the four parameters $(m_1,m_2,\theta_1,\theta_2)$ that are also displayed in Fig.~\ref{fig:GW-corner-ablation-studies}, as well as the pose $(t_{H}, t_{L})$. We compare chained NPE as described in section~\ref{sec:gwpe-results} to an oracle version: for the earlier the pose is inferred using standard NPE (green) while for the latter we take an oracle pose provided by a (slow) nested sampling algorithm (teal). We observe, that the result using the oracle pose matches the MCMC reference posterior well, while the other one shows clear deviations. Both versions use the same density estimator for the non-pose parameters $\phi\subset\theta$. This demonstrates that inaccuracies of the chained NPE baselines can be almost entirely attributed to inaccurate inital estimates of the pose. Poor pose estimates can occur since the density estimator trained to extract the pose operates on non pose-standardized data. 
    \\Note: The MCMC reference algorithm LALInference does not provide full pose information since it automatically marginalizes over $t_c$. For the oracle pose we thus employ the nested sampling algorithm bilby~\citep{Ashton:2018jfp}. 
    }
    \label{fig:GW-ablation-studies-GW170814-chained-pose}
\end{figure}

\begin{figure}
    \centering
    \begin{subfigure}{0.48\textwidth}
        \includegraphics[width=\textwidth]{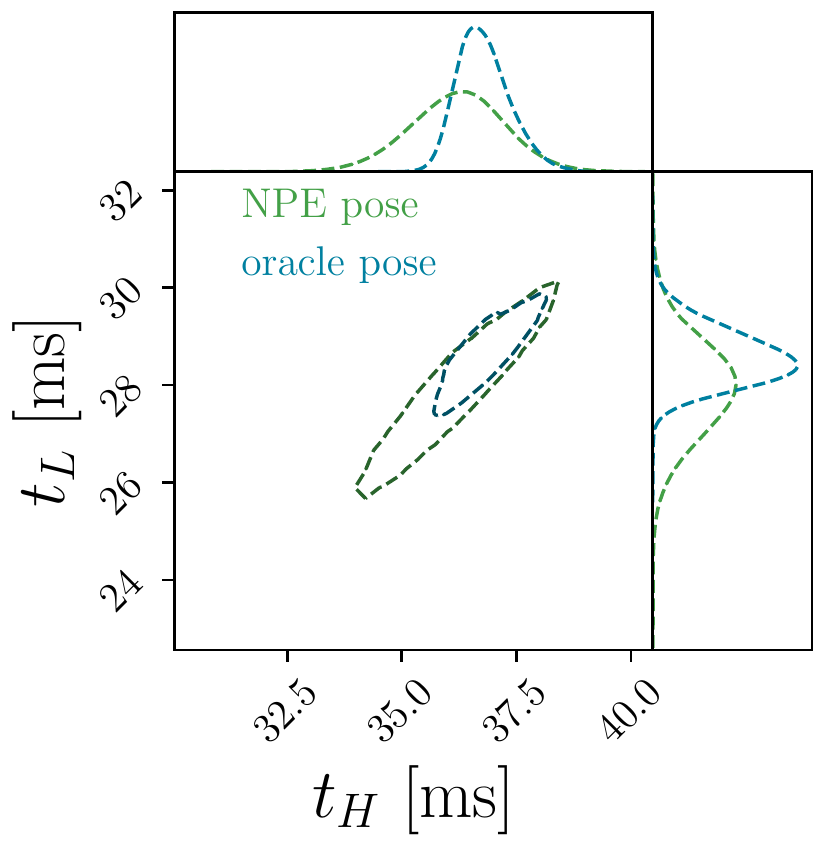}
    \end{subfigure}
    \hfill
    \begin{subfigure}{0.48\textwidth}
        \includegraphics[width=\textwidth]{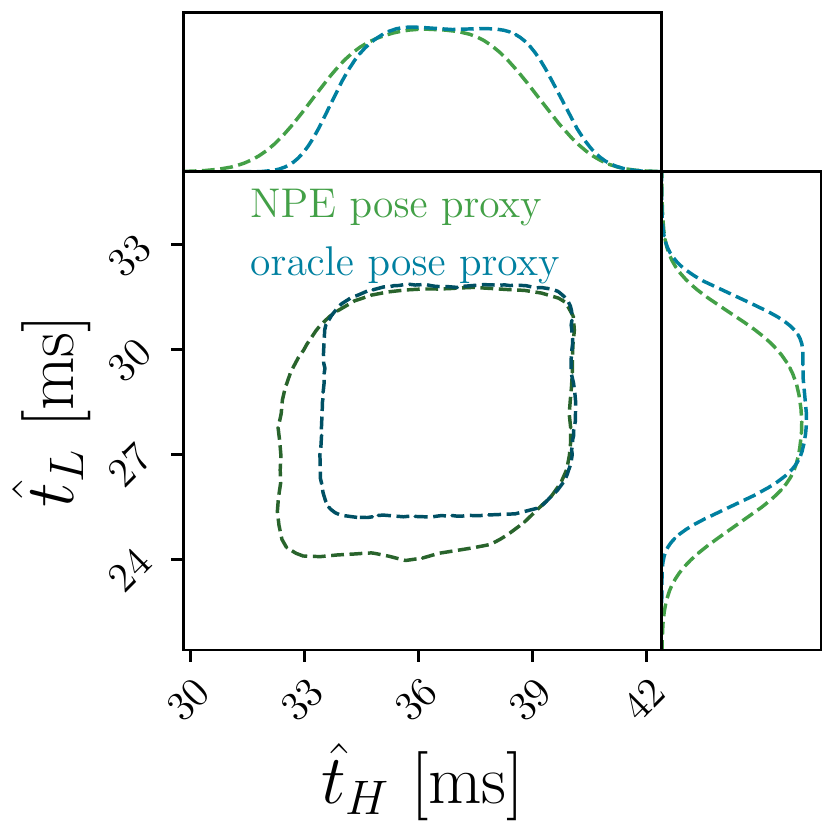}
    \end{subfigure}
    \caption{Left: Pose parameters $t_H$ and $t_L$ for the GW event GW170814, estimated with the neural density estimator $q_\text{init}$ with standard NPE (green), as well as the ``true'' pose inferred with bilby (teal). 
    Right: Pose proxies $\hat t_H$ and $\hat t_L$ for the wide kernel $\kappa_\text{wide} = U[-3~\mathrm{ms},3~\mathrm{ms}]^{n_I}$. These are obtained from the pose estimates in the left panel via a convolution with $\kappa_\text{wide}$. 
    We observe that the deviation between the oracle and the NPE estimate is substantially smaller for the pose proxy than for the pose itself due to the blurring operation. This leads to a better performance of fast-mode GNPE (with $\kappa_\text{wide}$ and only one iteration) compared to chained NPE in section~\ref{sec:gwpe-results}. 
      }
    \label{fig:GW-ablation-studies-GW170814-pose-and-proxy}
\end{figure}

\end{document}